\documentclass{article}

\usepackage[dvipsnames, svgnames, x11names]{xcolor}
\usepackage{colortbl}
\usepackage{microtype}
\usepackage{graphicx}
\usepackage{subfigure}
\usepackage{booktabs} 
\usepackage{multirow}
\usepackage{longtable}
\usepackage{hyperref}

\usepackage[accepted]{icml2024}

\usepackage{amsmath}
\usepackage{amssymb}
\usepackage{mathtools}
\usepackage{amsthm}

\usepackage{bm}

\newcommand{\Expect}{\mathbb E}

\newcommand{\sizeofsol}{{\ell}}
\newcommand{\Inst}{\mathcal{C}}
\newcommand{\instance}{x}
\newcommand{\Instance}{\mathcal{X}}
\newcommand{\inst}{c}

\newcommand{\cost}{D}
\newcommand{\dist}{d}

\newcommand{\graph}{\mathcal{G}}
\newcommand{\Edge}{\mathcal{E}}

\newcommand{\demand}{z}
\newcommand{\capacity}{R}

\newcommand{\emb}{\bm{E}}

\usepackage{etoolbox}
\newtoggle{final}
\toggletrue{final} 
\definecolor{darkgreen}{rgb}{0 0.6 0}

\newcommand{\zhs}[1]{\iftoggle{final}{#1}{{\color{darkgreen} #1}}}

\newcommand{\methodfullname}{Invariant Nested View Transformer}
\newcommand{\methodname}{INViT}
\newcommand{\methodintro}{\textbf{I}nvariant \textbf{N}ested \textbf{Vi}ew \textbf{T}ransformer (\textbf{\methodname})}

\usepackage[capitalize,noabbrev]{cleveref}

\theoremstyle{plain}

\theoremstyle{definition}

\theoremstyle{remark}

\usepackage[textsize=tiny]{todonotes}

\makeatletter
\def\UrlAlphabet{%
      \do\a\do\b\do\c\do\d\do\e\do\f\do\g\do\h\do\i\do\j%
      \do\k\do\l\do\m\do\n\do\o\do\p\do\q\do\r\do\s\do\t%
      \do\u\do\v\do\w\do\x\do\y\do\z\do\A\do\B\do\C\do\D%
      \do\E\do\F\do\G\do\H\do\I\do\J\do\K\do\L\do\M\do\N%
      \do\O\do\P\do\Q\do\R\do\S\do\T\do\U\do\V\do\W\do\X%
      \do\Y\do\Z}
\def\UrlDigits{\do\1\do\2\do\3\do\4\do\5\do\6\do\7\do\8\do\9\do\0}
\g@addto@macro{\UrlBreaks}{\UrlOrds}
\g@addto@macro{\UrlBreaks}{\UrlAlphabet}
\g@addto@macro{\UrlBreaks}{\UrlDigits}
\makeatother

\icmltitlerunning{INViT: A Generalizable Routing Problem Solver with Invariant Nested View Transformer}

\begin{document}

\twocolumn[
\icmltitle{INViT: A Generalizable Routing Problem Solver \\ with Invariant Nested View Transformer}



\icmlsetsymbol{equal}{*}

\begin{icmlauthorlist}
\icmlauthor{Han Fang}{sjtu}
\icmlauthor{Zhihao Song}{sjtu}
\icmlauthor{Paul Weng}{dku}
\icmlauthor{Yutong Ban}{sjtu}
\end{icmlauthorlist}

\icmlaffiliation{sjtu}{Joint Institute of Michigan, Shanghai Jiao Tong University, Shanghai, China}
\icmlaffiliation{dku}{Duke Kunshan University, Jiangsu, China}

\icmlcorrespondingauthor{Paul Weng}{paul.weng@dukekunshan.edu.cn}
\icmlcorrespondingauthor{Yutong Ban}{yban@sjtu.edu.cn}

\icmlkeywords{Machine Learning, ICML}

\vskip 0.3in
]



\printAffiliationsAndNotice{}  

\begin{abstract}
Recently, deep reinforcement learning has shown promising results for learning fast heuristics to solve routing problems.
Meanwhile, most of the solvers suffer from generalizing to an unseen distribution or distributions with different scales.
To address this issue, we propose a novel architecture, called \methodintro{}, which is designed to enforce a nested design together with invariant views inside the encoders to promote the generalizability of the learned solver. 
It applies a modified policy gradient algorithm enhanced with data augmentations.
We demonstrate that the proposed \methodname{} achieves a dominant generalization performance on both TSP and CVRP problems with various distributions and different problem scales.
Code is avaiable at \href{https://github.com/Kasumigaoka-Utaha/INViT}{https://github.com/Kasumigaoka-Utaha/INViT}.
\end{abstract}

\section{Introduction}\label{sec:intro}

Among all combinatorial optimization problems, routing problems, such as traveling salesman problem (TSP) or vehicle routing problem (VRP), are arguably among the most studied thanks to their wide application range, such as logistics \citep{madani_balancing_2020}, electronic design automation \citep{alkaya_application_2013}, or bioinformatics \citep{matai_traveling_nodate}.
Due to their NP-hard nature, exact algorithms are impracticable for solving large-scale instances, which has motivated the active development of approximate heuristic methods.
Although state-of-the-art (SOTA) heuristic methods, such as LKH3 \citep{helsgaun_general_2009,helsgaun2017extension} or HGS \citep{Vidal_2022}, have been designed to provide high-quality solutions for large routing problem instances with higher efficiency, the computational costs remain prohibitively high.

\begin{figure}[tb]
    \centering
    \includegraphics[width=\linewidth]{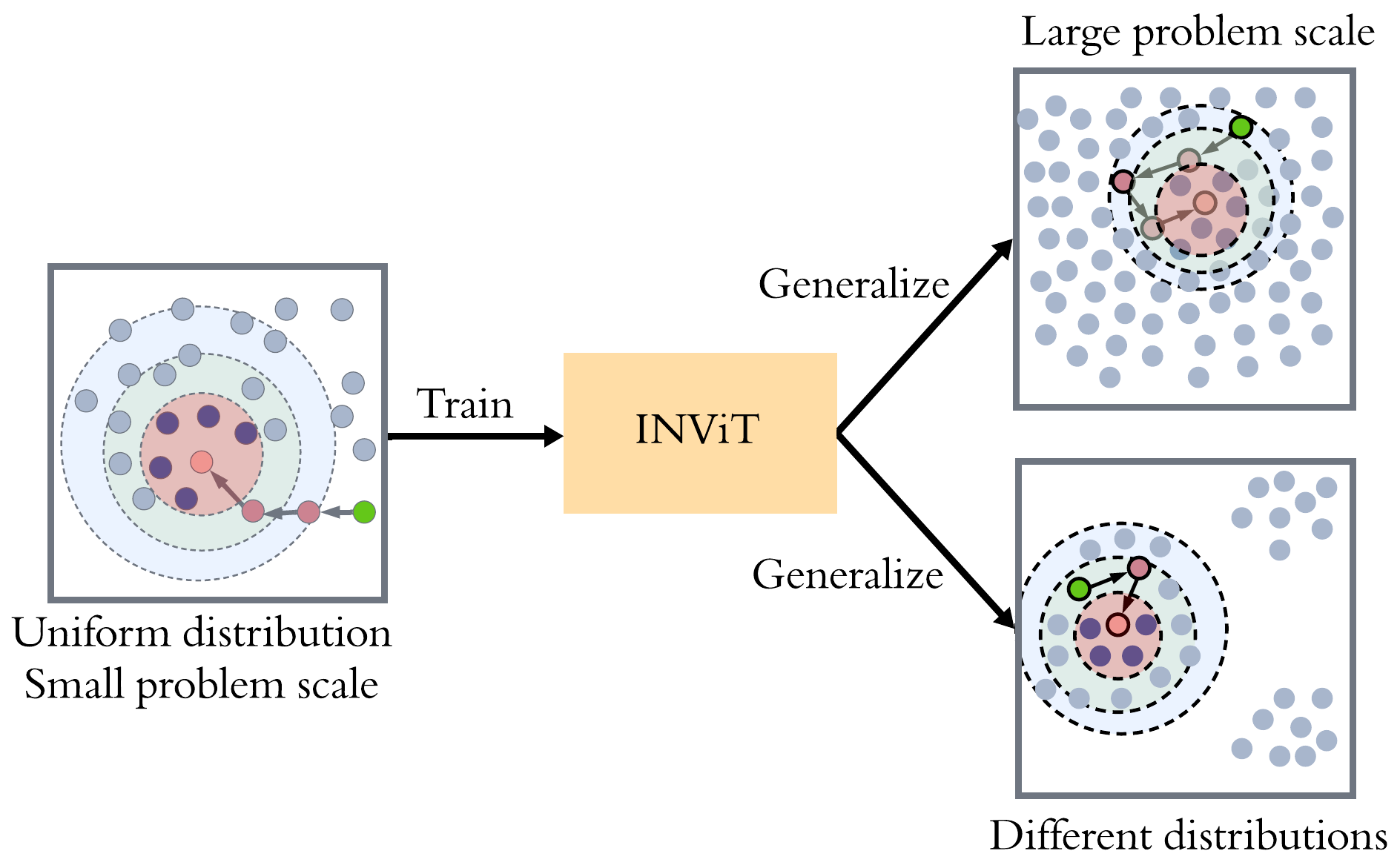}
    \vspace{-2ex}
    \caption{
             Our method \methodname{} aggregates node information from multiple nested local views (marked as colored discs). 
            Trained on small instances following uniform distributions, \methodname{} can generalize to instances with larger sizes or/and different distributions.} 
    \label{fig:teaser}
    \vspace{-3ex}
\end{figure}


To obtain faster heuristics, researchers have started to actively explore the exploitation of deep learning, and especially deep reinforcement learning (DRL), either (1) to learn to construct \citep{kool_attention_2018,jin_pointerformer_2023}, in which case the learned solver generates a solution step by step, or (2) to learn to search \citep{da_costa_learning_2021, fu_generalize_2021, Min_Bai_Gomes_2023, Falkner_Schmidt-Thieme_2023}, in which case the learned solver guides a local search method. 
In this paper, we focus on neural constructive methods, which usually enjoy faster inference while still reaching good performance compared with learn-to-search methods.

While constructive solvers demonstrate promising results, the existing DRL-based models generally lack robust generalization abilities, as also previously noted by \citet{Joshi_Cappart_Rousseau_Laurent_Bresson_2020}.
Indeed, those models are usually trained on fixed-scale (e.g., small) instances drawn from a fixed (e.g., uniform) probability distribution, but, once trained, they are incapable of generating satisfactory solutions on new instances of larger scales (i.e., cross-size generalization) or drawn from a different distribution (i.e., cross-distribution generalization).
While one may think of training on more diverse instances (larger scales or drawn from more diverse distributions) to address this generalization issue, this comes with an increased computational cost, which may be completely impractical for huge instances.
Moreover, after deployment, new instances with larger scales or drawn from unseen distributions may always happen.

In this paper, our objective is to develop a constructive (a.k.a. \emph{autoregressive}) solver with strong generalization capabilities, ensuring stable performance irrespective of distribution or scale, while also maintaining low time and memory complexity.
To that aim, we analyze previous models and identify two main sources for the generalization issue:
\emph{embedding aliasing} and \emph{interference from irrelevant nodes}.
The first source describes the situation where trained neural models fail to distinguish nodes in higher-density regions of a routing problem instance, as this \zhs{usually} happens when increasing instance sizes or when drawing from a non-uniform distribution.
The second source happens in particular in Transformer-based models where the self-attention weights take into account all the nodes, even the farthest ones, which are usually not relevant when constructing a solution.
As a countermeasure against these two phenomena, we propose the removal of nodes far from the last visited one in both the action and state spaces, which we justify by careful statistical analyses of optimal solutions in routing problems.

Motivated by our previous observations, we propose \methodintro{}, which combines graph sparsification and invariance, to address the generalization issue (see~\Cref{fig:teaser}). 
More specifically, \methodname{} is a Transformer-based architecture that processes multiple nested local views centered around the last visited node, where the smallest view only includes the most promising candidate actions, while the other larger views provide the most relevant state information for action selection.

Our contributions can be summarized as follows:
\begin{itemize}
    \item We identify two factors explaining the generalization issue observed in most previous DRL-based methods: embedding aliasing and interference from irrelevant nodes.
    By analyzing some statistical properties of optimal solutions of routing problems, we motivate the reduction of the state and action spaces.
    \item We design a novel Transformer-based architecture that takes invariant nested views of a routing problem instance. 
    Its architecture is justified by our previous observations and statistical analyses.
    \item We demonstrate on different datasets that the proposed architecture outperforms the current SOTA methods in terms of generalization on both TSP and CVRP.
\end{itemize}

\section{Related Work} \label{sec:related}

Recently, research investigating the application of deep learning and DRL to solve combinatorial optimization problems has become very active, exploring both local search and constructive methods.
For space reasons, we focus our discussion on the most related work for neural constructive methods.
In this literature, both novel architectures and novel DRL training algorithms have been proposed.
Our work mainly contributes in the first direction.

\paragraph{Architectures.}

Initial work in this direction proposed and studied various architectures, such as
Pointer Network \citep{vinyals_pointer_2015}, 
Attention Model \citep{kool_attention_2018}, and
Graph Neural Network (GNN) \citep{joshi_efficient_2019}. 
Apart from the latter one based on supervised learning, most studies consider reinforcement learning (RL), usually resorting to the simple REINFORCE algorithm \citep{williams_simple_1992}.
To the best of our knowledge, S2V-DQN \citep{khalil_learning_2017}, a sequence-to-vector architecture, trained by DQN \citep{Mnih_Kavukcuoglu_Silver_Graves_Antonoglou_Wierstra_Riedmiller_2013}, is the first method to explicitly consider cross-size generalization. 


The Attention Model \citep{kool_attention_2018} is based on the Transformer architecture \citep{Transformer}.
Given its generic nature and its promising performance, recent work has focused on improving its architecture.
For instance, PointerFormer \citep{jin_pointerformer_2023} develops a multi-pointer network to achieve better performance.
MVGCL \citep{Jiang_Cao_Wu_Zhang_2023} combines a GNN encoder followed by an attention-based encoder, where the former is trained by contrastive learning to leverage graph information for cross-distribution generalization.
\zhs{LEHD \citep{NEURIPS2023_LEHD} designs a heavy decoder to dynamically capture node features of varying input sizes for cross-size generalization.}
ELG\footnote{As a preprint on arXiv at the time of our submission, ELG refers to ELG-v1.}
proposed very recently \citep{Gao_Shang_Xue_Li_Qian_2023}, is an ensemble model comprised of a global policy and local policies, whose outputs are aggregated with a pre-fixed rule.
The local policies, utilizing k-nearest neighbors (k-NN) promote cross-size generalizability.

While ELG and our architecture share some superficial similarities (e.g., use of k-NN to create local views), there are some key differences, which make our proposition superior in terms of generalization performance. 
For instance, our method learns to aggregate the local views in the embedding space.
This directly tackles embedding aliasing, which is further reduced by considering nested local views in our method.
In addition, our architecture does not include a global view, since the use of a global encoder may be detrimental to the overall performance, as suggested by our statistical analysis (see~\Cref{sec:generalize}).

\paragraph{Training Algorithms.}
Most work applies standard RL algorithms, but some recent propositions specifically aim at improving the training (and also inference) algorithm.
For instance, 
POMO \citep{kwon_pomo_2020} includes a generic technique, which can significantly enhance the performance of neural solvers with minimal additional costs:
it generates multiple solutions by considering shifted starting nodes or by applying invariant transformations to input instances. 
Like other recent works \citep{jin_pointerformer_2023, Gao_Shang_Xue_Li_Qian_2023}, we also apply this simple but effective idea.
%

Regarding the generalization issue, some works aim at improving cross-size generalization, e.g., using meta RL \citep{Qiu_Sun_Yang_2022}, exploiting equivariance and local search \citep{ouyang_improving_2021, ouyang_generalization_2021}, or developing combinatorial problems as bisimulation quotiented Markov Decision Process \citep{NEURIPS2023_BQ-NCO}.
Others focus on cross-distribution generalization, e.g., using a specifically-designed new loss \citep{Jiang_Wu_Cao_Zhang_2022} or via knowledge distillation \citep{Bi_Ma_Wang_Cao_Chen_Sun_Chee_2022}.
Recently, Omni-TSP/VRP \citep{Zhou_Wu_Song_Cao_Zhang_2023}, inspired by the meta-RL approach proposed by \citet{Qiu_Sun_Yang_2022}, tackles both cross-size and cross-distribution generalization, as we do in our work.
The latter work is therefore a good SOTA baseline to compare with our proposition.

\begin{figure}[t!]
    \centering
    \includegraphics[width=1.0\linewidth]{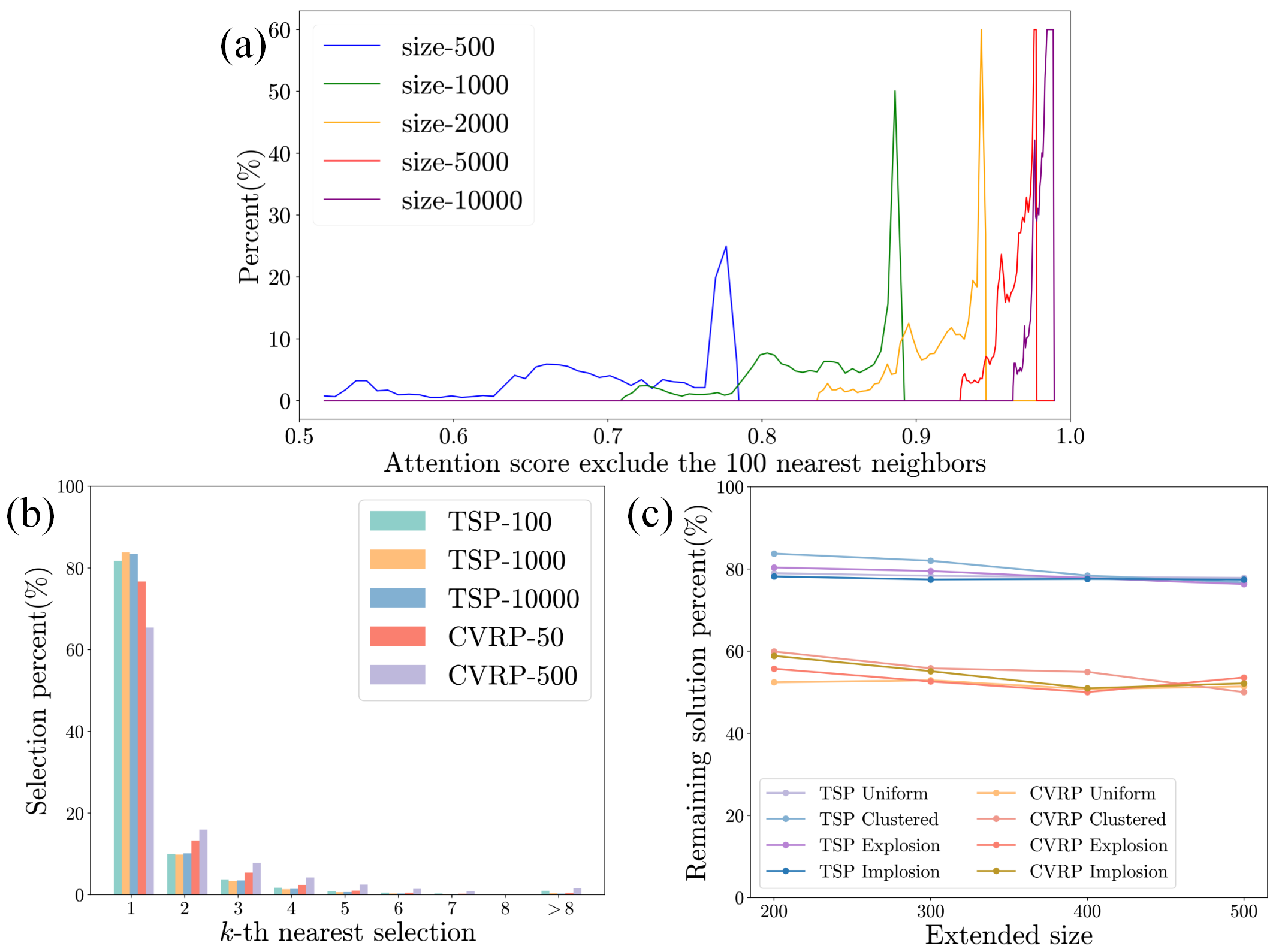}
    \vspace{-4ex}
    \caption{Preliminary findings. (a) The histogram of attention score for farther nodes in attention-based encoders (trained on TSP/CVRP 100). (b) The histogram of the optimal solution in $k$-th nearest neighbors for different $k$. (c) The percentage overlap of optimal solutions between the original and augmented instances.}
    \vspace{-3.5ex}
    \label{fig:motivation}
\end{figure}

\section{Background and Motivation}\label{sec:back}

Before introducing the motivation of \methodname{}, we first recall the basic formulation of routing problems and the common mechanism of autoregressive solvers.
Then based on our preliminary experiments, we point out the potential problems on generalization of the autoregressive solvers and propose some initial ideas to address those problems.

\subsection{Autoregressive Solvers for Routing Problems}\label{sec:pf}
\paragraph{Euclidean Routing Problem.}
Assume we are given a Euclidean Routing Problem instance $\instance$ with a graph $\graph$ and a set of constraints.
The graph $\graph$ is composed of a node set $\Inst = \{ \inst_1, \inst_2, \ldots, \inst_n \}$ and an edge set $\Edge$ contains all connections between nodes.
A feasible solution $F_{\instance}$ is an index sequence $(f_1, \cdots, f_\sizeofsol)$ of length $\sizeofsol$ that satisfies all the constraints.
Basically, each node $\inst_i$ has a coordinate $(c_i^1, c_i^2) \in [0, 1]^2$.
The cost is defined by $\cost_\instance({F}_{\instance}) = \dist(\inst_{f_\sizeofsol}, \inst_{f_1}) + \sum_{t = 1}^{\sizeofsol-1} \dist(\inst_{f_t}, \inst_{f_{t+1}}) \,,$ where $\dist$ is the Euclidean distance.
Our goal is to find a feasible solution that minimizes the cost function.
The constraints vary according to the specific Routing Problem.
For TSP, the only constraint is that the agent has to visit all the nodes exactly once.
For VRP, an extra set of variable, demands, is introduced to constrain the behavior of the agent.
Each node $\inst_i$ has a demand $\demand_i$ to fulfill and the agent has a fixed capacity $\capacity$.
A depot node $\inst_0$ is introduced for the agent to replenish when it runs out of its capacity.
In Capacitated VRP (CVRP), the agent is constrained to visit nodes except depot strictly once.

\paragraph{Autoregressive Solvers.}
Such a solver starts from an initial node, and repeatedly selects the next node to visit, until it outputs a feasible solution.
Regarding this process as a Markov decision process (MDP), at time step $t$, a state $s_t$ consists of a partial solution $ (\inst_{f_1}, \cdots, \inst_{f_t})$ and a remaining graph $\graph_t=\left( \Inst_t, \Edge_t \right)$. 
As noticed by \citet{kool_attention_2018}, the stateful partial solution can be reduced to the first visited node/depot and the last visited node $(\inst_{f_1}, \inst_{f_t})$.
A solver calculates a probability $p_\theta(a_t|s_t)$ for each action $a_t$ (i.e., node to be visited at time $t$) given the observable state $s_t$.
By the chain rule, the joint probability of a feasible solution is given by:
\begin{equation}
p_\theta \left( {F}_{\instance}|\instance \right)=\prod_{t=1}^{\sizeofsol}p_\theta(a_t|s_t).
\end{equation}
The REINFORCE \citep{williams_simple_1992} algorithm can train an autoregressive solver using gradient $\nabla_\theta \mathcal{L} (\pi_\theta)$ defined by:
\begin{equation}
    \Expect_{p_\theta}\left[ \left( \cost_\instance \left( {F}_{\instance}\right) - b (\instance) \right) \nabla_\theta \log \prod_{t=1}^{\sizeofsol}p_\theta(a_t|s_t) \right],
    \label{eq:reinforce}
\end{equation}
where $b (\instance)$ represents a baseline performance.

\subsection{Generalization Issue in Embedding Space}\label{sec:generalize}

To design an autoregressive solver that can generalize well both in the cross-size and cross-distribution settings, we first identify two shortcomings of current neural solvers trained on small scale and uniform distributed instances: \emph{embedding aliasing} and \emph{interference from irrelevant nodes}.

\paragraph{Embedding Aliasing.}
Recall that (deep) neural networks are simply Lipschitz functions \citep{ScamanVirmaux18}.
Let $h$ denotes the encoder layer of a neural solver trained on uniformly-distributed instances in the unit square, with size bounded by $n$.
Then, $h$ would satisfy the following Lipschitz inequality:
\begin{equation}
    \left\|h(\inst_i)-h(\inst_j)\right\|\leq L\left\|\inst_i-\inst_j\right\|,\qquad \forall \inst_i,\inst_j\in\Inst
\end{equation}
where $L>0$ is the Lipschitz constant.

After training, encoder $h$ should be able to usually distinguish nodes whose expected minimum pairwise distance in the unit square is $O(1/n)$.
However, when considering a new uniformly-distributed instance whose size $N$ is larger than $n$, encoder $h$ would have to distinguish nodes whose expected minimum pairwise distance is $O(1/N) < O(1/n)$.
Because of the Lipschitz inequality, there will be necessarily a size $N$ such that the embeddings produced by encoder $h$ will be mixed up, leading to incorrect action choices by the neural solver.

We call this phenomenon embedding aliasing, which provides one partial explanation to the generalization issues observed in existing neural solvers.
Note that embedding aliasing can also occur when considering different distributions. 
Indeed, a non-uniform distribution will necessarily generate some regions that contain densely packed nodes.

\paragraph{Interference from Irrelevant Nodes.} 
The second issue mostly impacts attention-based solvers that process the complete graph directly.
Recall that one attention layer computes the following embeddings:
\begin{equation}
    \emb = \texttt{Softmax}\left(\frac{QK^\top}{\sqrt{d_K}}\right)V.
\end{equation}
where $Q, K, V$ correspond to query, key and value. 
For node $c_j$, its impact on the embedding of node $c_i$ is given by the attention score between $Q_i$ and $K_j$. 
After training, the encoder may learn to assign lower attention scores to irrelevant nodes (i.e., far away nodes in routing problems), however, the cumulative impact of those irrelevant nodes becomes non-negligible as the instance size increases, as illustrated in \cref{fig:motivation} (a), which shows the empirical distribution of the sum of attention scores after excluding the 100 closest neighbors for a given model trained on instance size 100.
Mechanically, the contribution of those irrelevant nodes may impact the new embeddings, further amplifying the embedding aliasing issue.

\begin{figure*}[htbp]
    \centering
    \includegraphics[width=\linewidth]{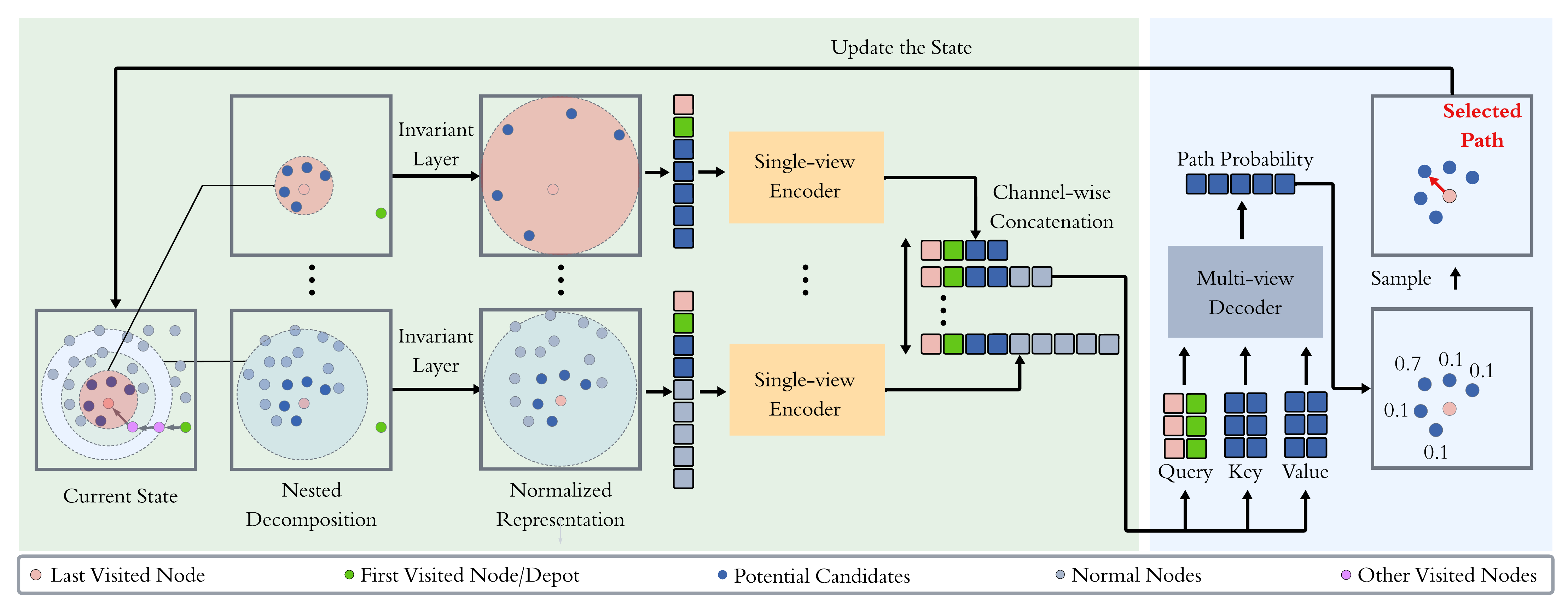}
    \vspace{-4ex}
    \caption{The overall architecture of \methodname{}.
             The input state is extracted into multiple nested views, consisting of neighborhood nodes around the last visited node. 
             Nodes located in the smallest view are potential candidates, and other nodes located in the view are called normal nodes.
             Each nested view is processed by a single-view encoder to obtain the embeddings for each node. Embeddings are then concatenated channel-wisely  across different views.
             The decoder takes the embeddings of the last visited node and the first visited node (or depot) as the query, and the embeddings of the potential candidates as the key and the value.
             Lastly, the model samples a node to visit by the output probabilities. 
             It updates the partial tour in an autoregressive manner until a complete tour is constructed.}
    \label{fig:archi}
\end{figure*}

\subsection{Preliminary Findings}\label{sec:obs}

The previous observations suggest to control the number of nodes given as inputs of a neural solver.
Interestingly, both the action and state spaces could be reduced.

\paragraph{Action Space.}
While theoretically, the action space should contain all nodes that satisfy the constraints of a problem (e.g., unvisited node in TSP or unvisited node whose demand is less than the current capacity in CVRP).
In practice, only the closest nodes need to be considered, as justified by \cref{fig:motivation} (b), which shows the distribution of the rank in terms of neareast neighbors for any node in an optimal solution both in random TSP and CVRP with different scales.
This observation, which is quite natural since an optimal solution minimizes a sum of distances, 
indicates that we can reduce our action space to a smaller subset, composed of the closest neighbors of the last visited node within the original action space (e.g., 8-NN can include $>$98\% of optimal choices).

\paragraph{State Space.}
While the action space can be narrowed down, the action choice can still depend on eliminated nodes.
Indeed, they could provide useful information regarding the future impact.
To simulate node elimination, we instead add randomly-distributed nodes outside the unit square for different random instances.
\cref{fig:motivation} (c) measures the percentage of edges that appear both in the optimal solutions of the initial random instances and in those of the augmented instances.
These results indicate that the eliminated actions have a relatively limited impact on the optimal choices for TSP.
In CVRP, the impact is larger, due to the capacity constraint, which can result in larger changes of the optimal solution.
However, since the effects are not as pronounced as for the action space and because of the two issues described in \Cref{sec:generalize}, a state space reduction with several nested sets may be beneficial.

\section{Method}\label{sec:archi}
We present \methodintro{}. We address the problems in \cref{sec:generalize} by implementing the observations in \cref{sec:obs} into the model design.



\subsection{\methodfullname{}}

The overall architecture and the autoregressive workflow are shown in \cref{fig:archi}.
\methodname{} incorporates a collection of nested-view encoders to embed the node features and maintains invariant views irrespective of the distribution and scale. 
According to \cref{sec:generalize}, we define a set of neighbors of the last visited node as potential candidate set $\mathbb{A}^p$, which corresponds to the smallest view in \cref{fig:archi}.
It's noteworthy that in VRP, the depot is typically considered as a candidate, since the agent is only prohibited from revisiting the depot when it is currently located there.

\paragraph{Nested View Encoders.}
As illustrated in \cref{sec:generalize}, the complete graph assumption allows farther nodes to have an exaggerated impact on the embeddings.
To tackle this issue, one simple approach is to perform sparsification on the graph.
Calculating a sparse graph for a static graph is a feasible task. 
However, taking into account the dynamic nature of the routing problem, computing a dynamically sparse graph during the inference procedure becomes a computationally expensive task.
Hence, we present a nested view encoder design to tackle this issue by sparsifying the graph into subgraphs, each composed of different numbers of neighbors.
As the k-nearest neighbor (k-NN) algorithm can offer stable neighbors and operate in a batch manner, we employ it to perform graph sparsification.
By eliminating the nodes which are not in neighbors with different k, multiple subgraphs are produced.
After proceeding with the invariant layer, each parallel single-view encoder would receive a distinct invariant subgraph and output the embeddings under different graph-contexts. 
The nested view design enables \methodname{} to integrate the embedding with different correlations, emphasizing the correlations between highly related nodes while preserving some correlations between less related nodes.

\paragraph{Invariant Layer.}
As shown in \cref{sec:generalize}, most of the encoders struggle to distinguish close nodes when the distance between nodes becomes sufficiently small. It is another key factor that hurts the generalization capability of attention-based encoders.
Therefore, to overcome the problem, we designed a layer called \emph{Invariant Layer}.
The \emph{Invariant Layer} consists of two steps: normalization and projection.
The normalization could be formulated as follows:
\begin{equation}
\hat{c}^m_i= \frac{c^m_i-\min\limits_{j\in\mathbb{N}} c^m_j}{\max\limits_{m\in\left\{1,2\right\}} \max\limits_{i,j\in\mathbb{N}}|c_i^m-c_j^m|}.
\label{eq:norm}
\end{equation}
As shown in \cref{fig:archi}, in addition to the potential candidate set $\mathbb{A}^p$ and the last visited node, the subgraph also includes the first visited node (or depot), the impact of which cannot be neglected.
However, it is possible the first visited node (or depot) falls outside the region of the potential candidate set $\mathbb{A}^p$ and the last visited node, potentially compromising the effectiveness of the normalization process.
In cases where the first visited node (or depot) cannot be visited, we incorporate a projection step, which could be formulated as
\begin{equation}
\hat{c}^m_0= \texttt{clip}\left( \frac{c^m_0-\min\limits_{j\in\mathbb{N}} c^m_j}{\max\limits_{m\in\left\{1,2\right\}} \max\limits_{i,j\in\mathbb{N}}|c_i^m-c_j^m|},0,1\right),
\label{eq:proj}
\end{equation}
where $\texttt{clip}(u, v, w)=\max(v,\min(u,w))$ projects the out-of-region first visited node (or depot) to the boundary, ensuring an invariant boundary for its coordinate.

\paragraph{Single-view Encoder.}
Following the nested view design, multiple independent encoders are constructed to embed node features for different subgraphs.
Once processed by the \emph{invariant layer}, the subgraph is then input to the respective \emph{single-view encoder}.
The \emph{single-view encoder} is composed of an initial linear layer and several encoder blocks consisting of Multi-Head Attention modules and Feed Forward layers \citep{Transformer}.
To note that, the encoder does not use the positional encoding module, as the order of the input sequence is irrelevant to the Routing Problem.
In alignment with the design of the invariant layer, which encompasses normalization and projection on different nodes, distinct initial linear layers are applied to capture node features.
For each input subgraph, its node embeddings are produced by the corresponding \emph{single-view encoder}.

\paragraph{Multi-view Decoder.}
\methodname{} aggregates multiple single-view embeddings by a channel-wise concatenation then inputs to a \emph{multi-view decoder} after processing subgraphs by parallel single-view encoders.
As mentioned previously, each subgraph shares a common intersection of nodes, which is the potential candidate set $\mathbb{A}^p$, the last visited nodes, and the first visited nodes (or depot).
Discarding all the embeddings for those nodes outside this intersection, we can construct multi-view embeddings by channel-wisely concatenating single-view embeddings.
The embeddings of the last visited node and the first visited node (or depot) are input as the query for the \emph{decoder}, while the other embeddings of potential candidates serve as the key and the value in the \emph{decoder}.
The final output probability is the attention weight of the last layer, which could be formulated as
\begin{equation}
    \pi_t(a)=\left\{
    \begin{array}{lr}
    \texttt{Softmax}\left(C\cdot\tanh\left(\frac{Q\cdot K^T}{\sqrt{d_K}}\right)\right),& a\in \mathbb{A}^p \\
    0,& \text{Otherwise} \\
    \end{array}
    \right.
\end{equation}
where $C$ is a positive constant. With the probability map, the next node could be sampled, and the entire model operates in an autoregressive manner.

\subsection{Algorithm}\label{sec:algo}

\paragraph{Training Stage.}
To train \methodname{}, we apply a REINFORCE-based algorithm integrated with data augmentation.
Initially, two identical models are initialized with training parameter $\theta$ and baseline parameter $\theta^\texttt{BL}$.
At the training stage, random instances are generated for training.
For each training instance $x_i$, a set of augmented instances $\mathcal{X}_i=\{ \instance_{i,j}\}_{j=1}^{\omega}$ is generated using the augmentation function, which includes rotation, reflection, and normalization.
Leveraging the idea of \citet{kwon_pomo_2020}, we also introduce variation in the starting point for each augmented instance.
According to \cref{eq:reinforce}, the loss is computed based on the model performance and baseline performance.
Following \citet{kool_attention_2018}, the baseline tour is determined using the baseline model with a greedy rollout, while the model tour is computed using the training model with a random sampling strategy.
The performance is then calculated as follows:
\begin{equation}
\begin{aligned}
    &b(\instance_i) = \Expect_{\instance \in \Instance_i} \left[ \cost_{\instance} \left( {F}_{\instance} (\mu_{\theta^\texttt{BL}}) \right) \right] ,\\
    &\cost_{\instance_i}({F}_{\instance_i}) = \Expect_{\instance \in \Instance_i} \left[ \cost_{\instance} \left( {F}_{\instance} (\pi_{\theta}) \right) \right],\\
\end{aligned}
\end{equation}
where ${F}_{\instance} (\pi)$ denotes the tour of instance $\instance$ under policy $\pi$, and $\mu$ is the deterministic policy induced by $\pi$.

\paragraph{Test Stage.}
At the test stage, we also employ data augmentation to enhance the overall performance. 
In contrast to the training stage, where we aim to use the average performance to increase generalizability on augmented instances, we only seek the best solution at the test stage.
A comparison between the baseline model and training model is conducted at the end of each training epoch. 
If the training model outperforms the baseline model, its parameters $\theta^\texttt{BL}$ are substituted using the training parameter $\theta$.

\section{Experimental Results}\label{sec:res}

To validate the generalizability of the proposed \methodname{}, we use a series of datasets across various scales and distributions.
We also include a comprehensive evaluation of our method, with several SOTA baselines, on our generated datasets and well-known public datasets.

\begin{table*}[ht]
    \centering
    \setlength\tabcolsep{2pt}
    \caption{Performance on TSP problems with different distributions and problem scales. (* denotes imitation learning based methods and the rest are REINFORCE-based methods.)}\label{tab:main_res_1}
    \scriptsize
    \begin{tabular}{@{}c|cccccccc|cccccccc@{}}
    \toprule
    
     \textbf{Distribution} &\multicolumn{8}{c|}{\textbf{Uniform}} & \multicolumn{8}{c}{\textbf{Clustered}}\\
    \midrule
      Category & \multicolumn{2}{c}{TSP-100} & \multicolumn{2}{c}{TSP-1000} & \multicolumn{2}{c}{TSP-5000} & \multicolumn{2}{c|}{TSP-10000} & \multicolumn{2}{c}{TSP-100} & \multicolumn{2}{c}{TSP-1000} & \multicolumn{2}{c}{TSP-5000} & \multicolumn{2}{c}{TSP-10000}\\
     Measurements  & gap(\%) &time(s) & gap(\%) &time(s) & gap(\%) &time(s) & gap(\%) &time(s) & gap(\%) &time(s) & gap(\%) &time(s) & gap(\%) & time(s)& gap(\%) & time(s) \\
     \midrule
     (Near-)Optimality & $0.00$ & $23.8$m & $0.00$ & $16.3$h & $0.00$ & $2.1$h & $0.00$ & $1.4$d & $0.00$ & $34.4$m & $0.00$ & $16.9$h & $0.00$ & $4.6$h & $0.00$ & $1.7$d \\
     \midrule
     POMO(NeurIPS-20) & $1.29$ & $2.0$m & $49.57$ & $1.6$m & $61.81$ & $1.2$m & $84.99$ & $3.8$m & $3.89$ & $1.7$m & $50.36$ & $1.6$m & $83.46$ & $1.2$m & $102.50$ & $3.9$m \\
     PointerFormer(AAAI-23) & $\bm{0.43}$ & $1.7$m & $34.29$ & $1.6$m & $39.61$ & $1.1$m & $71.85$ & $3.9$m & $3.96$ & $1.7$m & $43.89$ & $1.6$m & $60.18$ & $1.1$m & $103.70$ & $3.9$m \\
     Omni-TSP(ICML-23) & $2.55$ & $2.0$m & $20.25$ & $1.9$m & $50.30$ & $1.4$m & $62.56$ & $4.2$m & $3.62$ & $2.0$m & $23.13$ & $1.9$m & $57.74$ & $1.3$m & $71.47$ & $4.2$m \\
     ELG-v1 & $0.51$ & $3.2$m & $11.81$ & $3.4$m & $19.53$ & $1.9$m & $20.83$ & $10.1$m & $3.69$ & $3.2$m & $17.67$ & $3.0$m & $32.00$ & $1.9$m & $38.94$ & $10.2$m \\
     \midrule
     *LEHD(NeurIPS-23) & $0.57$ & $11.5$m & $\bm{2.76}$ & $12.0$m & $15.80$ & $23.6$m & $24.10$ & $2.7$h & $4.51$ & $14.9$m & $13.74$ & $11.8$m & $35.70$ & $23.9$m & $54.50$ & $2.7$h \\
     *BQ-NCO(NeurIPS-23) & $5.90$ & $16.6$m & $3.91$ & $23.6$m & $12.70$ & $2.0$h & $18.78$ & $14.1$h & $8.86$ & $17.4$m & $19.17$ & $23.6$m & $53.72$ & $2.0$h & $89.40$ & $14.1$h \\
     
     \midrule
    \textbf{INViT-2V} & $1.65$ &$3.0$m & $6.15$ & $3.5$m & $6.88$ & $1.7$m & $6.18$ & $7.7$m & $3.12$ & $2.9$m & $9.32$ &$3.3$m & $9.07$ & $1.7$m & $9.02$ & $7.6$m \\
    \textbf{INViT-3V} & $0.95$ &$4.2$m & $5.99$ & $4.8$m & $\bm{6.46}$ & $2.2$m & $\bm{6.01}$ & $10.3$m & $\bm{2.47}$ & $4.0$m & $\bm{8.63}$ &$4.8$m & $\bm{8.57}$ & $2.2$m & $\bm{8.79}$ & $10.5$m \\
     \midrule
     \textbf{Distribution} &\multicolumn{8}{c|}{\textbf{Explosion}} & \multicolumn{8}{c}{\textbf{Implosion}}\\
    \midrule
      Category & \multicolumn{2}{c}{TSP-100} & \multicolumn{2}{c}{TSP-1000} & \multicolumn{2}{c}{TSP-5000} & \multicolumn{2}{c|}{TSP-10000} & \multicolumn{2}{c}{TSP-100} & \multicolumn{2}{c}{TSP-1000} & \multicolumn{2}{c}{TSP-5000} & \multicolumn{2}{c}{TSP-10000}\\
     Measurements  & gap(\%) &time(s) & gap(\%) &time(s) & gap(\%) &time(s) & gap(\%) &time(s) & gap (\%) & times(s) & gap(\%) &time(s) & gap(\%) &time(s)& gap(\%) & time(s) \\
    \midrule
     (Near-)Optimality & $0.00$ & $28.3$m & $0.00$ & $17.5$h & $0.00$ & $2.0$h & $0.00$ & $1.4$d & $0.00$ & $28.7$m & $0.00$ & $17.5$h & $0.00$ & $3.5$h & $0.00$ & $1.4$d \\
     \midrule
     POMO(NeurIPS-20) & $1.42$ & $1.7$m & $50.00$ & $1.6$m & $89.43$ & $1.2$m & $92.66$ & $3.8$m & $1.44$ & $1.7$m & $50.01$ & $1.6$m & $64.12$ & $1.2$m & $88.02$ & $3.8$m \\
     PointerFormer(AAAI-23) & $0.87$ & $1.7$m & $40.99$ & $1.6$m & $61.42$ & $1.1$m & $104.85$ & $3.8$m & $\bm{0.71}$ & $1.7$m & $35.21$ & $1.6$m & $39.97$ & $1.1$m & $74.57$ & $3.9$m \\
     Omni-TSP(ICML-23) & $3.21$ & $2.0$m & $21.97$ & $1.9$m & $54.45$ & $1.4$m & $66.72$ & $4.2$m & $2.67$ & $2.0$m & $20.29$ & $1.9$m & $52.01$ & $1.4$m & $63.85$ & $4.2$m \\
     ELG-v1 & $0.93$ & $3.5$m & $15.32$ & $3.0$m & $33.80$ & $1.9$m & $35.03$ & $9.8$m & $0.85$ & $3.2$m & $12.10$ & $3.1$m & $19.59$ & $1.9$m & $21.69$ & $9.8$m \\
     \midrule
     *LEHD(NeurIPS-23) & $\bm{0.68}$ & $11.1$m & $\bm{5.99}$ & $11.9$m & $21.34$ & $23.9$m & $30.66$ & $2.7$h & $1.17$ & $18.3$m & $\bm{4.25}$ & $12.4$m & $17.67$ & $23.6$m & $26.46$ & $2.7$h \\
     *BQ-NCO(NeurIPS-23) & $6.41$ & $18.0$m & $7.21$ & $23.4$m & $29.48$ & $2.0$h & $51.67$ & $14.1$h & $6.40$ & $16.8$m & $5.43$ & $23.8$m & $16.63$ & $2.0$h & $25.50$ & $14.1$h \\
     
     \midrule
     \textbf{INViT-2V} & $1.85$ &$3.1$m & $9.11$ & $3.5$m & $9.92$ & $1.7$m & $9.32$ & $7.6$m & $1.95$ & $2.9$m & $6.63$ &$3.4$m & $7.63$ & $1.7$m & $6.78$ & $7.6$m \\
    \textbf{INViT-3V} & $1.12$ &$4.3$m & $8.57$ & $4.7$m & $\bm{9.43}$ & $2.2$m & $\bm{9.05}$ & $10.2$m & $1.21$ & $4.0$m & $6.35$ &$4.8$m & $\bm{7.41}$ & $2.2$m  & $\bm{6.21}$ & $10.4$m \\
    \bottomrule
\end{tabular}
\end{table*}

\begin{table*}[ht]
    \centering
    \caption{\centering Performance on CVRP problems with different distributions and problem scales. (* denotes imitation learning based methods and the rest are REINFORCE-based methods.)}\label{tab:main_res_2}
    \scriptsize
\begin{tabular}{c|cccccc|cccccc}
\toprule
     \textbf{Distribution} &\multicolumn{6}{c|}{\textbf{Uniform}} & \multicolumn{6}{c}{\textbf{Clustered}}\\
    \midrule
     Category & \multicolumn{2}{c}{CVRP-50} & \multicolumn{2}{c}{CVRP-500} & \multicolumn{2}{c|}{CVRP-5000} & \multicolumn{2}{c}{CVRP-50} & \multicolumn{2}{c}{CVRP-500} & \multicolumn{2}{c}{CVRP-5000}\\
     Measurements & gap(\%) &time(s) & gap(\%) &time(s) & gap(\%) &time(s) & gap(\%) &time(s) & gap(\%) &time(s) & gap(\%) &time(s)\\
    \midrule
    (Near-)Optimality & $0.00$ & $5.1$h & $0.00$ & $1.1$d & $0.00$ & $3.3$d & $0.00$ & $5.7$h & $0.00$ & $2.3$d & $0.00$ & $3.3$d \\
    \midrule
    POMO(NeurIPS-20) & $6.01$ & $1.5$m & $32.85$ & $1.4$m & $277.63$ & $1.8$m & $6.68$ & $1.5$m & $26.80$ & $1.4$m & $182.90$ & $1.9$m \\
    Omni-VRP(ICML-23) & $5.15$ & $1.8$m & $7.06$ & $1.6$m & $36.17$ & $2.0$m & $3.84$ & $1.8$m & $\bm{5.08}$ & $1.6$m & $13.52$ & $1.9$m \\
    ELG-v1 & $3.74$ & $2.5$m & $7.07$ & $2.4$m & $11.80$ & $3.4$m & $4.98$ & $2.9$m & $6.28$ & $2.4$m & $15.46$ & $3.5$m \\
    \midrule
    *LEHD(NeurIPS-23) & $6.72$ & $11.1$m & $5.84$ & $7.5$m & $10.59$ & $25.0$m & $6.52$ & $7.2$m & $7.29$ & $7.4$m & $23.65$ & $25.1$m \\
    *BQ-NCO(NeurIPS-23) & $12.64$ & $10.0$m & $\bm{4.72}$ & $10.5$m & $5.37$ & $2.0$h & $13.93$ & $10.4$m & $5.93$ & $11.2$m & $14.44$ & $2.0$h \\
    
    \midrule
    \textbf{INViT-2V} & $3.82$ &$2.6$m & $8.75$ & $1.9$m & $6.05$ & $2.8$m & $3.98$ & $2.3$m & $8.28$ &$1.8$m & $6.93$ & $2.8$m\\
    \textbf{INViT-3V} & $\bm{3.04}$ &$3.7$m & $7.89$ & $2.6$m & $\bm{5.32}$ & $4.0$m & $\bm{3.12}$ & $3.2$m & $7.68$ &$2.6$m & $\bm{6.09}$ & $4.1$m\\
    \midrule
    \textbf{Distribution} &\multicolumn{6}{c|}{\textbf{Explosion}} & \multicolumn{6}{c}{\textbf{Implosion}}\\
    \midrule
     Category & \multicolumn{2}{c}{CVRP-50} & \multicolumn{2}{c}{CVRP-500} & \multicolumn{2}{c|}{CVRP-5000} & \multicolumn{2}{c}{CVRP-50} & \multicolumn{2}{c}{CVRP-500} & \multicolumn{2}{c}{CVRP-5000}\\
     Measurements & gap(\%) &time(s) & gap(\%) &time(s) & gap(\%) &time(s) & gap(\%) &time(s) & gap(\%) &time(s) & gap(\%) &time(s)\\
    \midrule
    (Near-)Optimality & $0.00$ & $5.0$h & $0.00$ & $2.8$d & $0.00$ & $3.3$d & $0.00$ & $5.0$h & $0.00$ & $2.0$d & $0.00$ & $3.3$d \\
    \midrule
    POMO(NeurIPS-20) & $6.39$ & $1.5$m & $33.91$ & $1.4$m & $226.27$ & $1.8$m & $6.30$ & $1.5$m & $31.45$ & $1.3$m & $253.96$ & $1.8$m \\
    Omni-VRP(ICML-23) & $4.95$ & $1.7$m & $7.02$ & $1.6$m & $20.58$ & $1.9$m & $5.09$ & $1.8$m & $6.56$ & $1.6$m & $30.62$ & $1.9$m \\
    ELG-v1 & $3.94$ & $2.5$m & $7.06$ & $2.4$m & $12.56$ & $3.4$m & $\bm{3.82}$ & $2.5$m & $6.93$ & $2.4$m & $11.68$ & $3.4$m \\
    \midrule
    *LEHD(NeurIPS-23) & $6.65$ & $7.0$m & $6.64$ & $11.4$m & $18.05$ & $25.3$m & $7.03$ & $7.1$m & $7.16$ & $7.4$m & $19.79$ & $25.0$m \\
    *BQ-NCO(NeurIPS-23) & $12.96$ & $10.0$m & $\bm{5.79}$ & $10.7$m & $15.06$ & $2.0$h & $13.41$ & $10.4$m & $\bm{5.28}$ & $10.9$m & $6.60$ & $2.0$h \\
    \midrule
    \textbf{INViT-2V} & $4.05$ &$2.8$m & $8.99$ & $1.9$m & $7.11$ & $2.9$m & $4.41$ & $2.3$m & $8.73$ &$1.8$m & $6.12$ & $2.9$m\\
    \textbf{INViT-3V} & $\bm{3.78}$ &$3.8$m & $8.26$ & $2.7$m & $\bm{6.03}$ & $4.2$m & $\bm{3.82}$ & $3.5$m & $7.33$ &$2.6$m & $\bm{5.09}$ & $4.2$m\\
    \bottomrule
    \end{tabular}
\end{table*}

\subsection{Experimental Setups}\label{sec:setup}

\paragraph{MSVDRP Dataset.}
We have produced a dataset called Multi-Scale Various-Distribution Routing Problem (MSVDRP) dataset.
The dataset contains multiple subsets featuring both cross-distribution and cross-size instances for TSP and CVRP.
The data generation process follows \citet{Bossek_Kerschke_Neumann_Wagner_Neumann_Trautmann_2019}, yielding 16 subsets for TSP, encompassing 4 distributions (uniform, clustered, explosion, and implosion) and 4 scales (TSP-100, TSP-1000, TSP-5000 and TSP-10000).
Additionally, 12 subsets for CVRP are generated under the same distributions but at three scales (CVRP-50, CVRP-500, and CVRP-5000).
The number of instances for each subset varies according to the scale, with 2000 instances for TSP-100/CVRP-50, 200 instances for TSP-1000/CVRP-500, and 20 instances for TSP-5000/TSP-10000/CVRP-5000.


\paragraph{Public Datasets.}
Furthermore, we also use public datasets: TSPLIB and CVRPLIB to validate the performance. 
These instances have diverse problem scales and adhere to real-world distributions.
For TSP, we include all symmetric instances in TSPLIB95 \citep{reinelt1991tsplib} with nodes represented as Euclidean 2D coordinates, containing 77 instances varying in scale from 51 to 18512.
For CVRP, we include all instances in CVRPLIB Set-X by \citet{Uchoa_Pecin_Pessoa_Poggi_Vidal_Subramanian_2017}, containing 100 instances varying in scale from 100 to 1000.

\paragraph{Evaluation Metrics.} \label{sec:eval}
For each comparison method, we report the average gap to the (near-)optimal solutions, solved by Gurobi \citep{gurobi} (for TSP-100), LKH3 \citep{helsgaun_general_2009, helsgaun2017extension} (for TSP-1000, TSP-5000 and TSP-10000), HGS \citep{Vidal_2022} (for CVRP), or given optimality (for TSPLIB and CVRPLIB).
\zhs{Each gap corresponding to a problem instance $x$ is calculated as follows:
\begin{equation}
    \mathrm{gap} = \dfrac{\cost_\instance (F^\mathrm{model}) - \cost_\instance (F^\mathrm{opt})}{\cost_\instance (F^\mathrm{opt})} \times 100\%, 
\end{equation}
where $D_x (F^\mathrm{model})$ represents the length of the model solution, and $D_x (F^\mathrm{opt})$ represents the length of the (near-)optimal solutions.}
Note that LKH3 and HGS may not produce exactly optimal solutions, but the comparison between reported gaps can still be guaranteed to be fair due to utilizing the same evaluation instances.
We also report the total inference time on each dataset for each neural constructive method.

\paragraph{Comparison Methods.}
As mentioned in related work, numerous neural constructive works share similar objectives with ours. 
We choose to compare with representative baseline methods including SOTA methods delineated in \cref{sec:related}: \textbf{REINFORCE-based neural constructive methods}, including POMO \citep{kwon_pomo_2020}, Omni-TSP/VRP \citep{Zhou_Wu_Song_Cao_Zhang_2023}, PointerFormer \citep{jin_pointerformer_2023}, ELG-v1 \citep{Gao_Shang_Xue_Li_Qian_2023} and \zhs{\textbf{imitation learning based neural constructive methods}, including LEHD \citep{NEURIPS2023_LEHD}, BQ-NCO \citep{NEURIPS2023_BQ-NCO}}.
The selected comparison methods can show the results in the following aspects: 1) demonstrating the occurrence of generalization issues for neural attention-based solvers and 2) illustrating the impacts of different methods on generalizability improvements.


\paragraph{Experimental Settings.}
During training, all the models including POMO, PointerFormer, \zhs{LEHD, BQ-NCO,} ELG-v1, and the proposed \methodname{} were  trained on TSP/CVRP of size 100 and with uniform distribution, except for Omni-TSP/VRP, which is trained on sizes from 50 to 200 and diverse distributions.
For the comparison methods, we use the pre-trained models provided by the authors.
For the proposed \methodname{}, the initial learning rate is set to $10^{-4}$, with a weight decay of 0.01.
The model is trained for \zhs{$1.5 \times 10^{5}$} steps, with a batch size of 128, \zhs{taking about 5 days on both TSP and CVRP}.
To specify the variants of our model, we use \methodname{}-2V (resp. \methodname{}-3V) to denote the \methodname{} model comprised of two (resp. three) single-view encoders, with k-NN size of 35, 15 (resp. 50, 35, 15).
\zhs{To make the total number of trainable parameters comparable to other baselines, each encoder is designed to have 2 layers.
Therefore, INViT-2V (resp. INViT-3V) contains 4 (resp. 6) layers of encoders in total, similar to most of the included baselines.}

Evaluations are performed on our MSVDRP dataset and the public datasets.
Following \citet{kwon_pomo_2020}, each method generates multiple solutions for an input instance using greedy rollout.
The number of solutions (pomo-size) is limited to 100, in case of memory issues for large-scale datasets.
During the evaluation, parallelization is not explored, i.e., each iteration only contains one test instance.
All the experiments are performed on the same machine, equipped with a single Intel Core i7-12700 CPU and a single RTX 4090 GPU.
More detailed experimental settings can be found in \cref{sec:exp_d}.

\subsection{Performance Analysis}

\paragraph{Performances on the MSVDRP Datasets.}
\cref{tab:main_res_1} and \cref{tab:main_res_2} demonstrates the performance on the MSVDRP datasets.
It can be observed that POMO and PointerFormer have huge gap increases both from TSP-100 (resp. CVRP-100) to TSP-1000/TSP-5000/TSP10000 (resp. CVRP-500/CVRP5000), and from uniform distribution to other distributions, especially to the clustered distributions.
This illustrates the existence of generalization issues in the attention-based models.

Following the tables, \methodname{}-3V achieves the best results on all large-scale datasets (TSP-5000 with average gap $7.97\%$, TSP-10000 with average gap $7.52\%$ and CVRP-5000 with average gap $5.63\%$), showing its great cross-size generalizability. Except for BQ-NCO on CVRP-5000 instances, other baselines fail to achieve satisfactory performance (average gap $>10\%$) on these large-scale datasets.


According to the result table, the relative gap increase from uniform distribution to other distributions for INViT-3V for TSP-5000 and TSP-10000 (resp. CVRP-5000) is $31\%$ and $33\%$ (resp. $8\%$), only worse than $9\%$ and $8\%$ (resp. $-40\%$) by Omni-TSP/VRP.
Importantly, our method is only trained on uniform distributions, different from Omni-TSP/VRP.
As indicated by the results of POMO, cross-size imposes more generalization difficulties on the model than cross-distribution. We observe that our model also outperforms Omni-TSP/VRP on all large-scale datasets.
This indicates that our proposed method enjoys good cross-distribution generalizability while being cross-size generalizable.

It can also be observed that \methodname{}-2V has a similar generalization performance on TSP and CVRP with less inference time. Having an additional single-view encoder, \methodname{}-3V has a slight improvement in all experiments.

Nevertheless, our method does not outperform the comparison methods on \zhs{some small-scale datasets (e.g. TSP-100) and most of the medium-scale datasets (e.g. TSP-1000, CVRP-500) while the difference between the performance of INViT-3V and the best baseline on these two datasets are subtle. 
One possible explanation is that such scales are close to size of training instances.
In such cases, embedding aliasing does not fully take effect and interference from irrelevant nodes does not have much negative influence on the embedding.
This explanation also corresponds to the fact that our method performs much closer to the best baseline on TSP-1000 (relative gap $10\%$) than on CVRP-500 (relative gap $43\%$), since embedding aliasing has a strong effect on size 1000.}



\paragraph{Performances on Public Dataset.}

\begin{table}[ht]
    \centering
    \caption{Performances on TSPLIB and CVRPLIB problems.}
    \scriptsize
    \begin{tabular}{@{}ccccc@{}}
        \toprule
        \textbf{TSPLIB} &  \textbf{$1\sim 100$} & \textbf{$101\sim 1000$} &\textbf{$1001\sim 10000$} &\textbf{$>10000$} \\
        \midrule
        POMO & $1.92\%$ & $13.49\%$ & $60.05\%$ & $94.27\%$ \\
        PointerFormer & $1.36\%$ & $10.63\%$ & $30.38\%$ & $52.46\%$ \\
        Omni-TSP & $1.98\%$ & $5.06\%$ & $31.53\%$ & $82.92\%$ \\
        LEHD & $\bm{0.64}\%$ & $\bm{3.42}\%$ & $12.46\%$ & $43.61\%^*$ \\
        BQ-NCO & $8.66\%$ & $8.35\%$ & $14.50\%$ & $45.21\%^*$ \\
        ELG-v1 & $1.15\%$ & $7.72\%$ & $16.80\%$ & $26.73\%$ \\
        \midrule
        \textbf{INViT-2V} & $1.84\%$ & $4.77\%$ & $8.81\%$ & $9.54\%$ \\
        \textbf{INViT-3V} & $1.14\%$ & $4.26\%$ & $\bm{8.61\%}$ & $\bm{9.11\%}$ \\
        \bottomrule
         \\
        \toprule
        \multicolumn{2}{c}{\textbf{CVRPLIB Set-X}} &  \textbf{$100\sim 200$} & \textbf{$201\sim 500$} &\textbf{$>500$} \\
        \midrule
        \multicolumn{2}{c}{POMO} & $9.43\%$ & $19.76\%$ & $58.82\%$ \\
        \multicolumn{2}{c}{Omni-VRP} & $7.80\%$ & $\bm{8.18}\%$ & $11.21\%$ \\
        \multicolumn{2}{c}{LEHD} & $11.75\%$ & $9.45\%$ & $17.60\%$ \\
        \multicolumn{2}{c}{BQ-NCO} & $13.35\%$ & $10.11\%$ & $11.18\%$ \\
        \multicolumn{2}{c}{ELG-v1} & $6.77\%$ & $8.95\%$ & $12.21\%$ \\
        \midrule
        \multicolumn{2}{c}{\textbf{INViT-2V}} & $7.23\%$ & $9.58\%$ & $10.39\%$ \\
        \multicolumn{2}{c}{\textbf{INViT-3V}} & $\bm{6.52\%}$ & $9.11\%$ & $\bm{10.21\%}$ \\
        \bottomrule
    \end{tabular}
    \label{tab:lib}
\end{table}

We group the results of TSPLIB and CVRPLIB Set-X by size in \cref{tab:lib}.
A marked star (*) represents the occurrence of out-of-memory issues.
Detailed results are displayed in \cref{sec:lib_d}.
As a supplement to the MSVDRP dataset, the conclusion that our method \methodname{} has a strong generalization ability still holds in these public datasets.
Meanwhile, our method \methodname{} achieves a comparable performance with comparison methods on small-scale \zhs{and medium-scale} instances, which is better than the comparison on MSVDRP Datasets. 
This potentially benefits from our strong generalization ability on unseen distributions.

\subsection{Ablation Study}
\begin{table}[H]
    \centering
    \caption{Ablation study on architecture variants.}
    \scriptsize
    \begin{tabular}{@{}ccccc@{}}
        \toprule
        \textbf{Variants} &  \textbf{TSP-1000} & \textbf{TSP-10000} & \textbf{CVRP-500} & \textbf{CVRP-5000} \\
        \midrule
        INViT-2V (Global) & $10.90\%$ & $15.58\%$ & $12.95\%$ & $13.45\%$ \\
        INViT-2V (w/o Inv) & $26.32\%$ & $66.38\%$ & $17.32\%$ & $20.48\%$ \\
        INViT-2V ($n_\text{head}=4$)& $8.10\%$ & $8.41\%$ & $9.45\%$ & $7.35\%$ \\
        INViT-2V ($n_\text{aug}=4$)& $8.05\%$ & $8.13\%$ & $9.12\%$ & $7.01\%$ \\
        INViT-2V (w/o Aug) & $8.83\%$ & $8.93\%$ & $9.77\%$ & $7.55\%$ \\
        INViT-3V (w/o Aug) & $8.05\%$ & $8.28\%$ & $9.23\%$ & $6.76\%$ \\
        INViT-2V (Model-50)& $8.95\%$ & $9.47\%$ & $10.18\%$ & $9.00\%$ \\
        INViT-3V (Model-50)& $8.34\%$ & $8.85\%$ & $9.92\%$ & $8.55\%$ \\
        INViT-1V & $14.87\%$ & $15.14\%$ & $13.21\%$ & $12.10\%$ \\
        INViT-4V & $7.35\%$ & $7.64\%$ & $8.77\%$ & $5.87\%$ \\
        \midrule
        \textbf{INViT-2V}& $7.80\%$ & $8.01\%$ & $9.01\%$ & $6.76\%$ \\
        \textbf{INViT-3V}& $7.69\%$ & $7.86\%$ & $9.06\%$ & $6.01\%$ \\    
        \bottomrule
    \end{tabular}
    \vspace{-3ex}
    \label{tab:ablation}
\end{table}
\vspace{-1ex}

We have conducted several ablation experiments to demonstrate the impacts of different \zhs{model designs}.
\methodname{}-2V (Global), consists of \zhs{two} single-view encoders, but one of the encoders \zhs{process} the global information without graph sparsification.
\methodname{} 2V (w/o Invariance) \zhs{excludes} the invariant layers from \methodname{}. 
\zhs{\methodname{}-2V ($n_\text{head} = 4$) changes the number of MHA heads to 4 (originally 8).
\methodname{}-2V ($n_\text{aug} = 4$) changes the number of generated augmented instances to 4 (originally 8).
\methodname{}-2V/3V (w/o Aug) removes data augmentation during training procedures.}
\methodname{}-2V (Model-50) and \methodname{}-3V (Model-50) are trained on TSP-50/CVRP-50 instances.
\methodname{}-1V only includes one single-view encoder, and we record the best result for each evaluation from multiple models trained by different k-NN sizes, i.e., 50, 35, 15 included in our \methodname{}-3V.
\zhs{\methodname{}-4V is the extended model composed of four single-view encoders, each with k-NN size of 75, 50, 35, and 15.}
The results on partial MSVDRP datasets are presented in \cref{tab:ablation}.
\zhs{Each reported gap is averaged among all four distributions, so the data can represent performance for both cross-size and cross-distribution.}

The experiments show that key components in the proposed architecture: the graph sparsification of all encoders \zhs{(excluded by INViT-2V (Global))}, the invariant transformation \zhs{(excluded by INViT-2V (w/o Inv))}, and the nested view design \zhs{(excluded by INViT-1V)}, all impose a positive effect on cross-size generalization.
\zhs{The change of hyperparameters and the removal of data augmentation do not impose significant changes on the performance.
It can be concluded that the proposed INViT architecture is robust against different training parameters.}
Increasing the scale of training instances (from size 50 to size 100) does improve the overall performance, meanwhile, our model still achieves a good performance by training on smaller-scale instances on TSP-50/CVRP-50, which again demonstrates the generalization capability of our model.
\zhs{We observe a marginal improvement led by accumulating single-view encoders, but the simplest implementation (INViT-2V) already enjoys great cross-size and cross-distribution generalizability.}

\section{Conclusion}
We present \methodfullname{} (\methodname{}), an autoregressive routing problem solver that has strong generalization capabilities on instances with larger scales and different distributions, which only requires training on small-scale uniform instances. 
Experiments demonstrate that \methodname{} outperforms SOTA autoregressive solvers on large-scale and cross-distribution instances.

\section*{Acknowledgments}

This work has been supported in part by the program of National Natural Science Foundation of China (No. 62176154) and by Shanghai Magnolia Funding Pujiang Program (No. 23PJ1404400).

\section*{Impact Statement}
Routing Problems are fundamental in combinatorial optimization problems, given its NP-hard nature and wide application range.
Exact algorithms or heuristic methods are computational expensive on large instances, due to its NP-hard nature.
Therefore, it is meaningful that we develop an architecture that could train on instances with small scale and uniform distribution but provide near-optimal solutions on instances with large-scale and unseen distributions with limited time and memory complexity.
In addition to improving the related machine learning fields, the contribution can be further applied to industries like logistics, electronic design automation or bioinformatics.



\bibliography{main}

\begin{thebibliography}{44}
\providecommand{\natexlab}[1]{#1}
\providecommand{\url}[1]{\texttt{#1}}
\expandafter\ifx\csname urlstyle\endcsname\relax
  \providecommand{\doi}[1]{doi: #1}\else
  \providecommand{\doi}{doi: \begingroup \urlstyle{rm}\Url}\fi

\bibitem[Alkaya \& Duman(2013)Alkaya and Duman]{alkaya_application_2013}
Alkaya, A.~F. and Duman, E.
\newblock Application of sequence-dependent traveling salesman problem in
  printed circuit board assembly.
\newblock \emph{IEEE Transactions on Components, Packaging and Manufacturing
  Technology}, 3\penalty0 (6):\penalty0 1063--1076, 2013.
\newblock \doi{10.1109/TCPMT.2013.2252429}.
\newblock URL \url{https://ieeexplore.ieee.org/abstract/document/6504483}.

\bibitem[Bello et~al.(2017)Bello, Pham, Le, Norouzi, and
  Bengio]{Bello_Pham_Le_Norouzi_Bengio_2017}
Bello, I., Pham, H., Le, Q.~V., Norouzi, M., and Bengio, S.
\newblock Neural combinatorial optimization with reinforcement learning.
\newblock arXiv, 2017.
\newblock URL \url{https://arxiv.org/abs/1611.09940}.

\bibitem[Bi et~al.(2022)Bi, Ma, Wang, Cao, Chen, Sun, and
  Chee]{Bi_Ma_Wang_Cao_Chen_Sun_Chee_2022}
Bi, J., Ma, Y., Wang, J., Cao, Z., Chen, J., Sun, Y., and Chee, Y.~M.
\newblock Learning generalizable models for vehicle routing problems via
  knowledge distillation.
\newblock In Koyejo, S., Mohamed, S., Agarwal, A., Belgrave, D., Cho, K., and
  Oh, A. (eds.), \emph{Advances in Neural Information Processing Systems},
  volume~35, pp.\  31226--31238. Curran Associates, Inc., 2022.
\newblock URL
  \url{https://proceedings.neurips.cc/paper_files/paper/2022/file/ca70528fb11dc8086c6a623da9f3fee6-Paper-Conference.pdf}.

\bibitem[Bossek et~al.(2019)Bossek, Kerschke, Neumann, Wagner, Neumann, and
  Trautmann]{Bossek_Kerschke_Neumann_Wagner_Neumann_Trautmann_2019}
Bossek, J., Kerschke, P., Neumann, A., Wagner, M., Neumann, F., and Trautmann,
  H.
\newblock Evolving diverse tsp instances by means of novel and creative
  mutation operators.
\newblock In \emph{Proceedings of the 15th ACM/SIGEVO Conference on Foundations
  of Genetic Algorithms}, FOGA '19, pp.\  58–71. Association for Computing
  Machinery, 2019.
\newblock ISBN 9781450362542.
\newblock \doi{10.1145/3299904.3340307}.
\newblock URL \url{https://doi.org/10.1145/3299904.3340307}.

\bibitem[Choo et~al.(2022)Choo, Kwon, Kim, Jae, Hottung, Tierney, and
  Gwon]{Choo_Kwon_Kim_Jae_Hottung_Tierney_Gwon_2022}
Choo, J., Kwon, Y.-D., Kim, J., Jae, J., Hottung, A., Tierney, K., and Gwon, Y.
\newblock Simulation-guided beam search for neural combinatorial optimization.
\newblock In Koyejo, S., Mohamed, S., Agarwal, A., Belgrave, D., Cho, K., and
  Oh, A. (eds.), \emph{Advances in Neural Information Processing Systems},
  volume~35, pp.\  8760--8772. Curran Associates, Inc., 2022.
\newblock URL
  \url{https://proceedings.neurips.cc/paper_files/paper/2022/file/39b9b60f0d149eabd1fff2d7c7d5afc4-Paper-Conference.pdf}.

\bibitem[da~Costa et~al.(2021)da~Costa, Rhuggenaath, Zhang, Akcay, and
  Kaymak]{da_costa_learning_2021}
da~Costa, P., Rhuggenaath, J., Zhang, Y., Akcay, A., and Kaymak, U.
\newblock Learning 2-{Opt} {Heuristics} for {Routing} {Problems} via {Deep}
  {Reinforcement} {Learning}.
\newblock \emph{SN Computer Science}, 2\penalty0 (5):\penalty0 388, July 2021.
\newblock ISSN 2661-8907.
\newblock \doi{10.1007/s42979-021-00779-2}.
\newblock URL
  \url{https://link.springer.com/content/pdf/10.1007/s42979-021-00779-2.pdf}.

\bibitem[Drakulic et~al.(2023)Drakulic, Michel, Mai, Sors, and
  Andreoli]{NEURIPS2023_BQ-NCO}
Drakulic, D., Michel, S., Mai, F., Sors, A., and Andreoli, J.-M.
\newblock Bq-nco: Bisimulation quotienting for efficient neural combinatorial
  optimization.
\newblock In Oh, A., Neumann, T., Globerson, A., Saenko, K., Hardt, M., and
  Levine, S. (eds.), \emph{Advances in Neural Information Processing Systems},
  volume~36, pp.\  77416--77429. Curran Associates, Inc., 2023.
\newblock URL
  \url{https://proceedings.neurips.cc/paper_files/paper/2023/file/f445ba15f0f05c26e1d24f908ea78d60-Paper-Conference.pdf}.

\bibitem[Falkner \& Schmidt-Thieme(2023)Falkner and
  Schmidt-Thieme]{Falkner_Schmidt-Thieme_2023}
Falkner, J.~K. and Schmidt-Thieme, L.
\newblock Too big, so fail? -- enabling neural construction methods to solve
  large-scale routing problems.
\newblock arXiv, 2023.
\newblock URL \url{https://arxiv.org/abs/2309.17089}.

\bibitem[Fu et~al.(2021)Fu, Qiu, and Zha]{fu_generalize_2021}
Fu, Z.-H., Qiu, K.-B., and Zha, H.
\newblock Generalize a small pre-trained model to arbitrarily large tsp
  instances.
\newblock In \emph{Proceedings of the AAAI Conference on Artificial
  Intelligence}, volume~35, pp.\  7474--7482, 2021.
\newblock \doi{10.1609/aaai.v35i8.16916}.
\newblock URL \url{https://ojs.aaai.org/index.php/AAAI/article/view/16916}.

\bibitem[Gao et~al.(2023)Gao, Shang, Xue, Li, and
  Qian]{Gao_Shang_Xue_Li_Qian_2023}
Gao, C., Shang, H., Xue, K., Li, D., and Qian, C.
\newblock Towards generalizable neural solvers for vehicle routing problems via
  ensemble with transferrable local policy.
\newblock arXiv, 2023.
\newblock URL \url{https://arxiv.org/abs/2308.14104v1}.

\bibitem[{Gurobi Optimization, LLC}(2023)]{gurobi}
{Gurobi Optimization, LLC}.
\newblock {Gurobi Optimizer Reference Manual}, 2023.
\newblock URL \url{https://www.gurobi.com}.

\bibitem[Helsgaun(2009)]{helsgaun_general_2009}
Helsgaun, K.
\newblock General k-opt submoves for the {Lin}–{Kernighan} {TSP} heuristic.
\newblock \emph{Mathematical Programming Computation}, 1\penalty0 (2):\penalty0
  119--163, October 2009.
\newblock ISSN 1867-2957.
\newblock \doi{10.1007/s12532-009-0004-6}.
\newblock URL
  \url{https://link.springer.com/content/pdf/10.1007/s12532-009-0004-6.pdf}.

\bibitem[Helsgaun(2017)]{helsgaun2017extension}
Helsgaun, K.
\newblock An extension of the lin-kernighan-helsgaun tsp solver for constrained
  traveling salesman and vehicle routing problems.
\newblock Technical report, Roskilde University, 2017.
\newblock URL
  \url{http://www.akira.ruc.dk/~keld/research/LKH-3/LKH-3_REPORT.pdf}.

\bibitem[Hottung et~al.(2022)Hottung, Kwon, and
  Tierney]{Hottung_Kwon_Tierney_2021}
Hottung, A., Kwon, Y.-D., and Tierney, K.
\newblock Efficient active search for combinatorial optimization problems.
\newblock In \emph{International Conference on Learning Representations}, 2022.
\newblock URL \url{https://openreview.net/forum?id=nO5caZwFwYu}.

\bibitem[Hou et~al.(2023)Hou, Yang, Su, Wang, and
  Deng]{Hou_Yang_Su_Wang_Deng_2022}
Hou, Q., Yang, J., Su, Y., Wang, X., and Deng, Y.
\newblock Generalize learned heuristics to solve large-scale vehicle routing
  problems in real-time.
\newblock In \emph{The Eleventh International Conference on Learning
  Representations}, 2023.
\newblock URL \url{https://openreview.net/forum?id=6ZajpxqTlQ}.

\bibitem[Jiang et~al.(2022)Jiang, Wu, Cao, and Zhang]{Jiang_Wu_Cao_Zhang_2022}
Jiang, Y., Wu, Y., Cao, Z., and Zhang, J.
\newblock Learning to solve routing problems via distributionally robust
  optimization.
\newblock In \emph{Proceedings of the AAAI Conference on Artificial
  Intelligence}, volume~36, pp.\  9786--9794, 2022.
\newblock \doi{10.1609/aaai.v36i9.21214}.
\newblock URL \url{https://ojs.aaai.org/index.php/AAAI/article/view/21214}.

\bibitem[Jiang et~al.(2023)Jiang, Cao, Wu, and Zhang]{Jiang_Cao_Wu_Zhang_2023}
Jiang, Y., Cao, Z., Wu, Y., and Zhang, J.
\newblock Multi-view graph contrastive learning for solving vehicle routing
  problems.
\newblock In Evans, R.~J. and Shpitser, I. (eds.), \emph{Proceedings of the
  Thirty-Ninth Conference on Uncertainty in Artificial Intelligence}, volume
  216 of \emph{Proceedings of Machine Learning Research}, pp.\  984--994. PMLR,
  2023.
\newblock URL \url{https://proceedings.mlr.press/v216/jiang23a.html}.

\bibitem[Jin et~al.(2023)Jin, Ding, Pan, He, Zhao, Qin, Song, and
  Bian]{jin_pointerformer_2023}
Jin, Y., Ding, Y., Pan, X., He, K., Zhao, L., Qin, T., Song, L., and Bian, J.
\newblock Pointerformer: Deep reinforced multi-pointer transformer for the
  traveling salesman problem.
\newblock In \emph{Proceedings of the AAAI Conference on Artificial
  Intelligence}, volume~37, pp.\  8132--8140, 2023.
\newblock \doi{10.1609/aaai.v37i7.25982}.
\newblock URL \url{https://ojs.aaai.org/index.php/AAAI/article/view/25982}.

\bibitem[Joshi et~al.(2019)Joshi, Laurent, and Bresson]{joshi_efficient_2019}
Joshi, C.~K., Laurent, T., and Bresson, X.
\newblock An efficient graph convolutional network technique for the travelling
  salesman problem.
\newblock arXiv, 2019.
\newblock URL \url{https://arxiv.org/abs/1906.01227}.

\bibitem[Joshi et~al.(2022)Joshi, Cappart, Rousseau, and
  Laurent]{Joshi_Cappart_Rousseau_Laurent_Bresson_2020}
Joshi, C.~K., Cappart, Q., Rousseau, L.-M., and Laurent, T.
\newblock Learning the travelling salesperson problem requires rethinking
  generalization.
\newblock \emph{Constraints}, 27\penalty0 (1):\penalty0 70–98, April 2022.
\newblock ISSN 1572-9354.
\newblock \doi{10.1007/s10601-022-09327-y}.
\newblock URL \url{https://doi.org/10.1007/s10601-022-09327-y}.

\bibitem[Khalil et~al.(2017)Khalil, Dai, Zhang, Dilkina, and
  Song]{khalil_learning_2017}
Khalil, E., Dai, H., Zhang, Y., Dilkina, B., and Song, L.
\newblock Learning combinatorial optimization algorithms over graphs.
\newblock In Guyon, I., Luxburg, U.~V., Bengio, S., Wallach, H., Fergus, R.,
  Vishwanathan, S., and Garnett, R. (eds.), \emph{Advances in Neural
  Information Processing Systems}, volume~30. Curran Associates, Inc., 2017.
\newblock URL
  \url{https://proceedings.neurips.cc/paper_files/paper/2017/file/d9896106ca98d3d05b8cbdf4fd8b13a1-Paper.pdf}.

\bibitem[Kool et~al.(2019)Kool, van Hoof, and Welling]{kool_attention_2018}
Kool, W., van Hoof, H., and Welling, M.
\newblock Attention, learn to solve routing problems!
\newblock In \emph{International Conference on Learning Representations}, 2019.
\newblock URL \url{https://openreview.net/forum?id=ByxBFsRqYm}.

\bibitem[Kwon et~al.(2020)Kwon, Choo, Kim, Yoon, Gwon, and Min]{kwon_pomo_2020}
Kwon, Y.-D., Choo, J., Kim, B., Yoon, I., Gwon, Y., and Min, S.
\newblock Pomo: Policy optimization with multiple optima for reinforcement
  learning.
\newblock In Larochelle, H., Ranzato, M., Hadsell, R., Balcan, M., and Lin, H.
  (eds.), \emph{Advances in Neural Information Processing Systems}, volume~33,
  pp.\  21188--21198. Curran Associates, Inc., 2020.
\newblock URL
  \url{https://proceedings.neurips.cc/paper_files/paper/2020/file/f231f2107df69eab0a3862d50018a9b2-Paper.pdf}.

\bibitem[Luo et~al.(2023)Luo, Lin, Liu, Zhang, and Wang]{NEURIPS2023_LEHD}
Luo, F., Lin, X., Liu, F., Zhang, Q., and Wang, Z.
\newblock Neural combinatorial optimization with heavy decoder: Toward large
  scale generalization.
\newblock In Oh, A., Neumann, T., Globerson, A., Saenko, K., Hardt, M., and
  Levine, S. (eds.), \emph{Advances in Neural Information Processing Systems},
  volume~36, pp.\  8845--8864. Curran Associates, Inc., 2023.
\newblock URL
  \url{https://proceedings.neurips.cc/paper_files/paper/2023/file/1c10d0c087c14689628124bbc8fa69f6-Paper-Conference.pdf}.

\bibitem[Ma et~al.(2019)Ma, Ge, He, Thaker, and Drori]{ma_combinatorial_2019}
Ma, Q., Ge, S., He, D., Thaker, D., and Drori, I.
\newblock Combinatorial optimization by graph pointer networks and hierarchical
  reinforcement learning.
\newblock arXiv, 2019.
\newblock URL \url{https://arxiv.org/abs/1911.04936}.

\bibitem[Madani et~al.(2021)Madani, Batta, and Karwan]{madani_balancing_2020}
Madani, A., Batta, R., and Karwan, M.
\newblock The balancing traveling salesman problem: application to warehouse
  order picking.
\newblock \emph{TOP}, 29\penalty0 (2):\penalty0 442–469, 2021.
\newblock ISSN 1863-8279.
\newblock \doi{10.1007/s11750-020-00557-y}.
\newblock URL \url{https://doi.org/10.1007/s11750-020-00557-y}.

\bibitem[Matai et~al.(2010)Matai, Singh, and Mittal]{matai_traveling_nodate}
Matai, R., Singh, S., and Mittal, M.~L.
\newblock Traveling salesman problem: an overview of applications,
  formulations, and solution approaches.
\newblock In Davendra, D. (ed.), \emph{Traveling Salesman Problem, Theory and
  Applications}, chapter~1. IntechOpen, Rijeka, 2010.
\newblock \doi{10.5772/12909}.
\newblock URL \url{https://www.intechopen.com/chapters/12736}.

\bibitem[Min et~al.(2023)Min, Bai, and Gomes]{Min_Bai_Gomes_2023}
Min, Y., Bai, Y., and Gomes, C.~P.
\newblock Unsupervised learning for solving the travelling salesman problem.
\newblock In \emph{Thirty-seventh Conference on Neural Information Processing
  Systems}, 2023.
\newblock URL \url{https://openreview.net/forum?id=lAEc7aIW20}.

\bibitem[Mnih et~al.(2013)Mnih, Kavukcuoglu, Silver, Graves, Antonoglou,
  Wierstra, and
  Riedmiller]{Mnih_Kavukcuoglu_Silver_Graves_Antonoglou_Wierstra_Riedmiller_2013}
Mnih, V., Kavukcuoglu, K., Silver, D., Graves, A., Antonoglou, I., Wierstra,
  D., and Riedmiller, M.
\newblock Playing atari with deep reinforcement learning.
\newblock arXiv, 2013.
\newblock URL \url{https://arxiv.org/abs/1312.5602}.

\bibitem[Ouyang et~al.(2021{\natexlab{a}})Ouyang, Wang, Han, Jin, and
  Weng]{ouyang_improving_2021}
Ouyang, W., Wang, Y., Han, S., Jin, Z., and Weng, P.
\newblock Improving generalization of deep reinforcement learning-based tsp
  solvers.
\newblock In \emph{2021 IEEE Symposium Series on Computational Intelligence
  (SSCI)}, pp.\  01--08, 2021{\natexlab{a}}.
\newblock \doi{10.1109/SSCI50451.2021.9659970}.
\newblock URL \url{https://ieeexplore.ieee.org/document/9659970}.

\bibitem[Ouyang et~al.(2021{\natexlab{b}})Ouyang, Wang, Weng, and
  Han]{ouyang_generalization_2021}
Ouyang, W., Wang, Y., Weng, P., and Han, S.
\newblock Generalization in deep rl for tsp problems via equivariance and local
  search.
\newblock arXiv, 2021{\natexlab{b}}.
\newblock URL \url{https://arxiv.org/abs/2110.03595}.

\bibitem[Pan et~al.(2023)Pan, Jin, Ding, Feng, Zhao, Song, and
  Bian]{Pan_Jin_Ding_Feng_Zhao_Song_Bian_2023}
Pan, X., Jin, Y., Ding, Y., Feng, M., Zhao, L., Song, L., and Bian, J.
\newblock H-tsp: Hierarchically solving the large-scale traveling salesman
  problem.
\newblock In \emph{Proceedings of the AAAI Conference on Artificial
  Intelligence}, volume~37, pp.\  9345--9353, 2023.
\newblock \doi{10.1609/aaai.v37i8.26120}.
\newblock URL \url{https://ojs.aaai.org/index.php/AAAI/article/view/26120}.

\bibitem[Qiu et~al.(2022)Qiu, Sun, and Yang]{Qiu_Sun_Yang_2022}
Qiu, R., Sun, Z., and Yang, Y.
\newblock Dimes: A differentiable meta solver for combinatorial optimization
  problems.
\newblock In Koyejo, S., Mohamed, S., Agarwal, A., Belgrave, D., Cho, K., and
  Oh, A. (eds.), \emph{Advances in Neural Information Processing Systems},
  volume~35, pp.\  25531--25546. Curran Associates, Inc., 2022.
\newblock URL
  \url{https://proceedings.neurips.cc/paper_files/paper/2022/file/a3a7387e49f4de290c23beea2dfcdc75-Paper-Conference.pdf}.

\bibitem[Reinelt(1991)]{reinelt1991tsplib}
Reinelt, G.
\newblock Tsplib—a traveling salesman problem library.
\newblock \emph{ORSA journal on computing}, 3\penalty0 (4):\penalty0 376--384,
  1991.
\newblock URL \url{http://comopt.ifi.uni-heidelberg.de/software/TSPLIB95/}.

\bibitem[Sun \& Yang(2023)Sun and Yang]{sun_difusco_2023}
Sun, Z. and Yang, Y.
\newblock {DIFUSCO}: Graph-based diffusion solvers for combinatorial
  optimization.
\newblock In \emph{Thirty-seventh Conference on Neural Information Processing
  Systems}, 2023.
\newblock URL \url{https://openreview.net/forum?id=JV8Ff0lgVV}.

\bibitem[Uchoa et~al.(2017)Uchoa, Pecin, Pessoa, Poggi, Vidal, and
  Subramanian]{Uchoa_Pecin_Pessoa_Poggi_Vidal_Subramanian_2017}
Uchoa, E., Pecin, D., Pessoa, A., Poggi, M., Vidal, T., and Subramanian, A.
\newblock New benchmark instances for the capacitated vehicle routing problem.
\newblock \emph{European Journal of Operational Research}, 257\penalty0
  (3):\penalty0 845–858, 2017.
\newblock URL
  \url{https://ideas.repec.org/a/eee/ejores/v257y2017i3p845-858.html}.

\bibitem[Vaswani et~al.(2017)Vaswani, Shazeer, Parmar, Uszkoreit, Jones, Gomez,
  Kaiser, and Polosukhin]{Transformer}
Vaswani, A., Shazeer, N., Parmar, N., Uszkoreit, J., Jones, L., Gomez, A.~N.,
  Kaiser, L.~u., and Polosukhin, I.
\newblock Attention is all you need.
\newblock In Guyon, I., Luxburg, U.~V., Bengio, S., Wallach, H., Fergus, R.,
  Vishwanathan, S., and Garnett, R. (eds.), \emph{Advances in Neural
  Information Processing Systems}, volume~30. Curran Associates, Inc., 2017.
\newblock URL
  \url{https://proceedings.neurips.cc/paper_files/paper/2017/file/3f5ee243547dee91fbd053c1c4a845aa-Paper.pdf}.

\bibitem[Vidal(2022)]{Vidal_2022}
Vidal, T.
\newblock Hybrid genetic search for the cvrp: Open-source implementation and
  swap* neighborhood.
\newblock \emph{Computers \& Operations Research}, 140:\penalty0 105643, 2022.
\newblock ISSN 0305-0548.
\newblock \doi{https://doi.org/10.1016/j.cor.2021.105643}.
\newblock URL
  \url{https://www.sciencedirect.com/science/article/pii/S030505482100349X}.

\bibitem[Vinyals et~al.(2015)Vinyals, Fortunato, and
  Jaitly]{vinyals_pointer_2015}
Vinyals, O., Fortunato, M., and Jaitly, N.
\newblock Pointer networks.
\newblock In Cortes, C., Lawrence, N., Lee, D., Sugiyama, M., and Garnett, R.
  (eds.), \emph{Advances in Neural Information Processing Systems}, volume~28.
  Curran Associates, Inc., 2015.
\newblock URL
  \url{https://proceedings.neurips.cc/paper_files/paper/2015/file/29921001f2f04bd3baee84a12e98098f-Paper.pdf}.

\bibitem[Virmaux \& Scaman(2018)Virmaux and Scaman]{ScamanVirmaux18}
Virmaux, A. and Scaman, K.
\newblock Lipschitz regularity of deep neural networks: analysis and efficient
  estimation.
\newblock In Bengio, S., Wallach, H., Larochelle, H., Grauman, K.,
  Cesa-Bianchi, N., and Garnett, R. (eds.), \emph{Advances in Neural
  Information Processing Systems}, volume~31. Curran Associates, Inc., 2018.
\newblock URL
  \url{https://proceedings.neurips.cc/paper_files/paper/2018/file/d54e99a6c03704e95e6965532dec148b-Paper.pdf}.

\bibitem[Williams(1992)]{williams_simple_1992}
Williams, R.~J.
\newblock Simple statistical gradient-following algorithms for connectionist
  reinforcement learning.
\newblock \emph{Machine Learning}, 8\penalty0 (3):\penalty0 229--256, 1992.
\newblock ISSN 1573-0565.
\newblock \doi{10.1007/BF00992696}.
\newblock URL
  \url{https://link.springer.com/content/pdf/10.1007/BF00992696.pdf}.

\bibitem[Xin et~al.(2021)Xin, Song, Cao, and Zhang]{Xin_Song_Cao_Zhang_2021}
Xin, L., Song, W., Cao, Z., and Zhang, J.
\newblock Multi-decoder attention model with embedding glimpse for solving
  vehicle routing problems.
\newblock In \emph{Proceedings of the AAAI Conference on Artificial
  Intelligence}, volume~35, pp.\  12042--12049, 2021.
\newblock \doi{10.1609/aaai.v35i13.17430}.
\newblock URL \url{https://ojs.aaai.org/index.php/AAAI/article/view/17430}.

\bibitem[Ye et~al.(2023)Ye, Wang, Liang, Cao, Li, and
  Li]{Ye_Wang_Liang_Cao_Li_Li_2023}
Ye, H., Wang, J., Liang, H., Cao, Z., Li, Y., and Li, F.
\newblock Glop: Learning global partition and local construction for solving
  large-scale routing problems in real-time.
\newblock arXiv, 2023.
\newblock URL \url{https://arxiv.org/abs/2312.08224}.

\bibitem[Zhou et~al.(2023)Zhou, Wu, Song, Cao, and
  Zhang]{Zhou_Wu_Song_Cao_Zhang_2023}
Zhou, J., Wu, Y., Song, W., Cao, Z., and Zhang, J.
\newblock Towards omni-generalizable neural methods for vehicle routing
  problems.
\newblock In Krause, A., Brunskill, E., Cho, K., Engelhardt, B., Sabato, S.,
  and Scarlett, J. (eds.), \emph{Proceedings of the 40th International
  Conference on Machine Learning}, volume 202 of \emph{Proceedings of Machine
  Learning Research}, pp.\  42769--42789. PMLR, 2023.
\newblock URL \url{https://proceedings.mlr.press/v202/zhou23o.html}.

\end{thebibliography}
\bibliographystyle{icml2024}

\newpage
\appendix
\onecolumn



\section{Appendix}

\subsection{Detailed Analysis} \label{sec:stat_d}

\begin{figure}[htbp]
    \centering
    \includegraphics[width=0.8\linewidth]{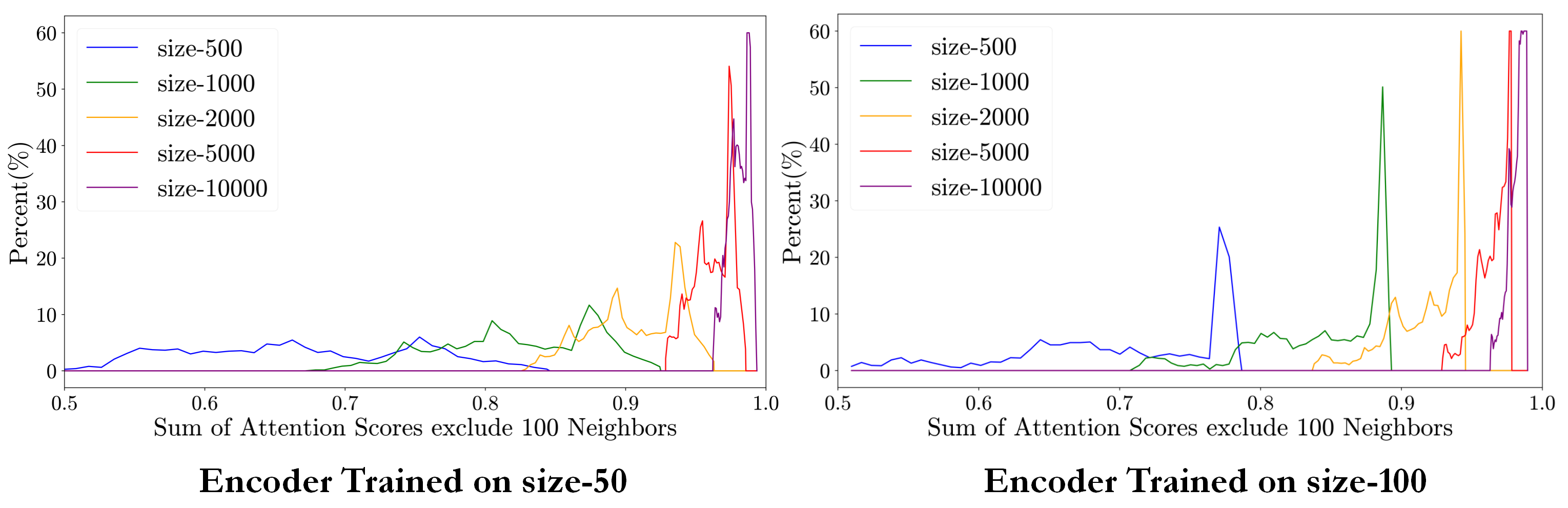}
    \caption{Impact of the nodes outside neighbor groups for the encoder.}
    \label{fig:att_score}
\end{figure}
\paragraph{Farther Node Impact for Encoder.}
To plot \cref{fig:att_score}, we train two standard attention-based solvers by uniform instances with scale-50/100, following POMO \citet{kwon_pomo_2020}.
Then we collect its attention scores between each node and calculate the cumulative sum of attention score outside 100 nearest neighbors.
\cref{fig:att_score} demonstrates the cumulative impact of farther nodes on the embedding.
It could be seen that with the increasing of scale, the peak of the distribution comes close to 1.
This means that neighbors actually have very limited effect on the embedding when encoding the whole graph simultaneously, which is opposite to our other observation.

\begin{figure}[htbp]
    \centering
    \includegraphics[width=0.7\linewidth]{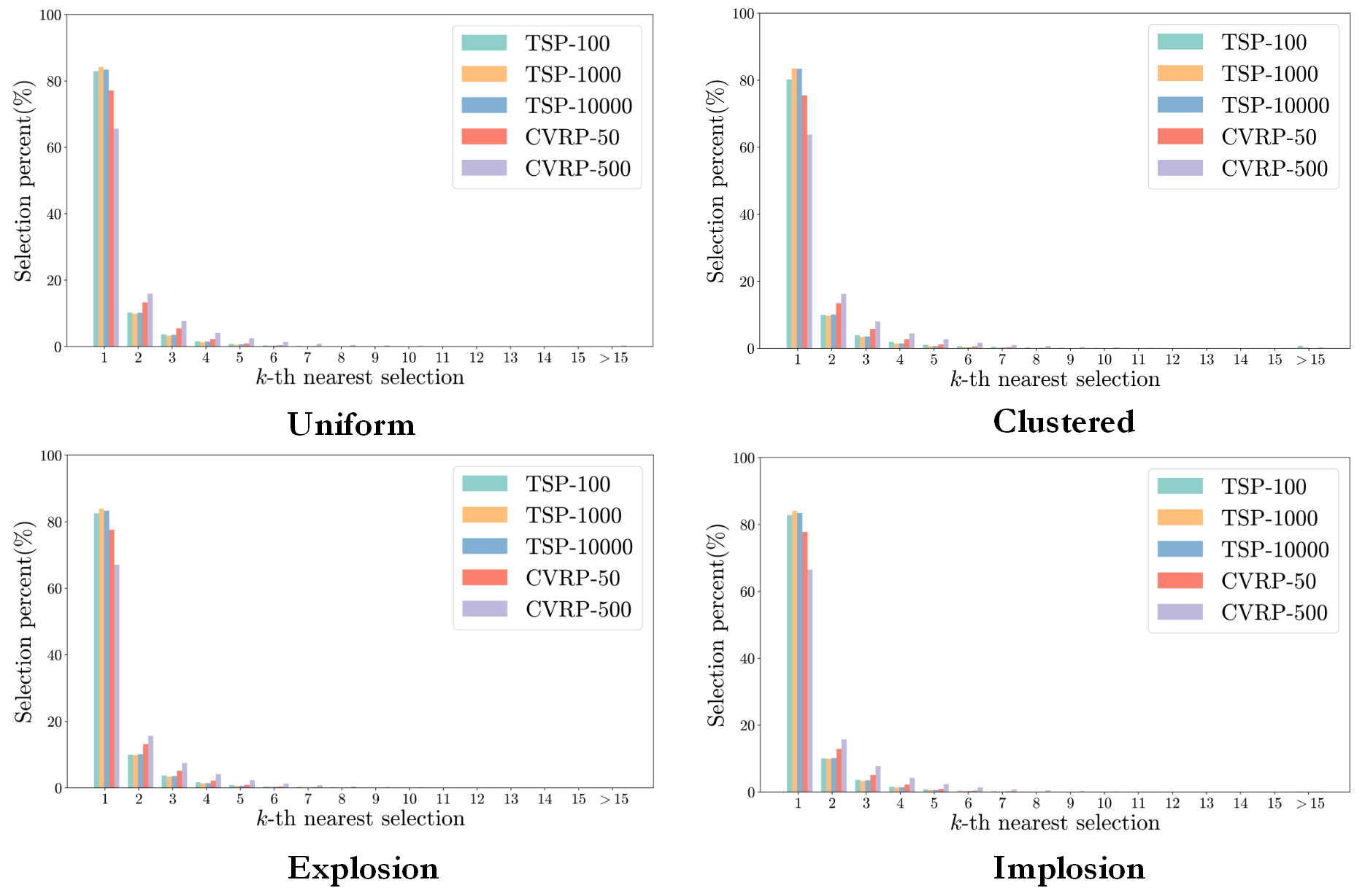}
    \caption{Statistics on the action choice of the optimal solution.
             It represents the distribution of the rank of the next node to visit from a node in a solution tour among the nearest neighbors of the latter.
             Best viewed in colors.}
    \label{fig:knn}
\end{figure}

\paragraph{K-NN Statistics.}
According to the cost function of TSP and CVRP, there is a natural tendency that the agent tend to select from the neighbors of the last visited node as its next visited node.
However, choosing from the neighbors is not always the best choice, there are cases such that the agent choose the node beyond the neighbors.
Therefore, we conduct a statistical analysis on the choice of the optimal tour for both TSP and CVRP under a constructive view.
The constructive view represents we emulate the inference procedure of auto-regressive solvers, which travels the tour step by step and only consider feasible actions.
As illustrated in \cref{fig:knn}, for both TSP and CVRP, there is a relatively low probability that the agent would select nodes outside the k-th (i.e., 15) neighbor irrespective of scale and distribution.

\begin{figure}[htbp]
    \centering
    \includegraphics[width=0.8\linewidth]{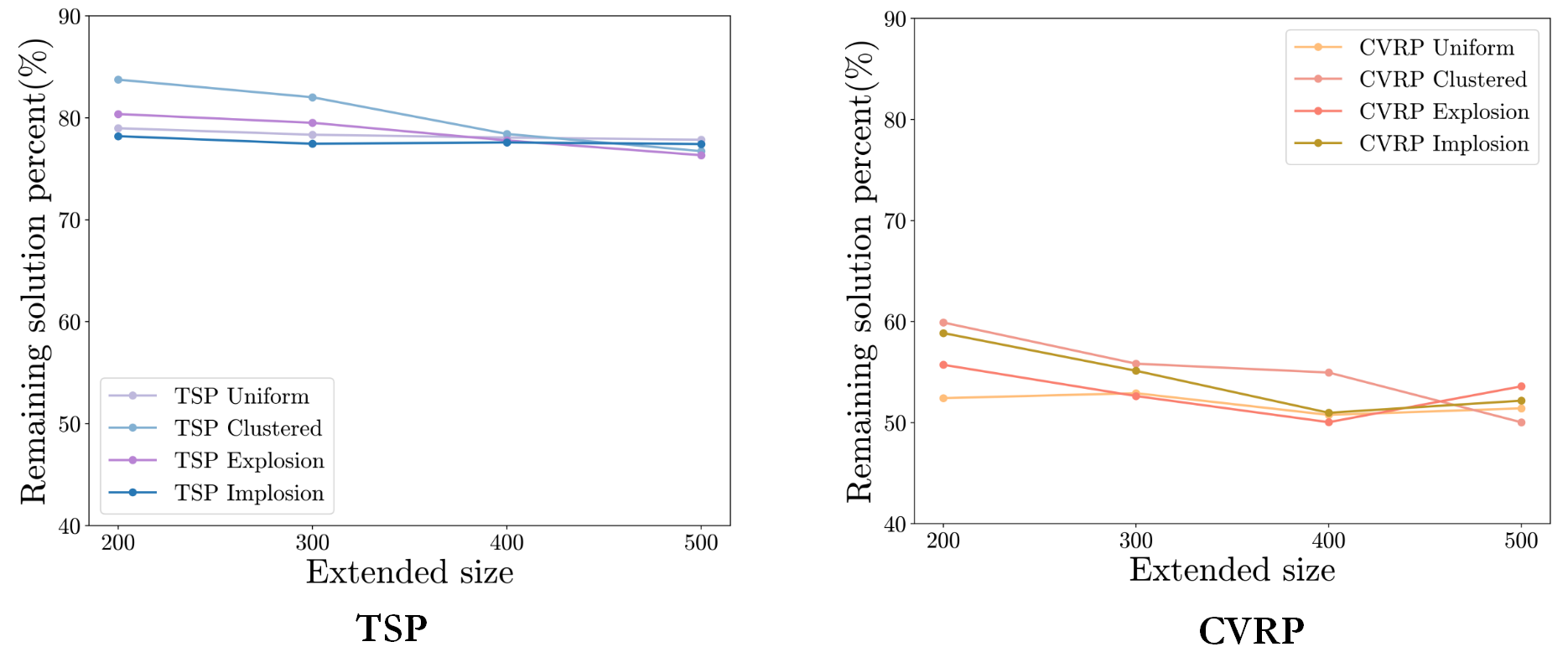}
    \caption{Optimal solution change curve.
             It represents the empirical solution change from small instances with a size of 100 to a larger size.
             Best viewed in colors.}
    \label{fig:change}
\end{figure}

\paragraph{Farther Node Impact for Optimal Solution.}
Though farther nodes might not be chosen as the next visited node, it might still have impact on the choice of the next visited node.
Therefore, to further investigate the real impact brought by farther nodes, we have done another statistical analysis on the optimal solutions.
We first generate TSP/CVRP instances with different distributions on the unit board and then calculate the optimal tours.
Then by adding nodes outside unit board, we get larger perturbed instances and also calculate the optimal tours.
Therefore, by calculating the percent of the edges remain from original solutions to the perturbed solutions, we are able to see whether the local behavior of the agent is dramatically changed by the additional nodes outside the unit board, which represents the farther nodes.
\cref{fig:change} demonstrates that for TSP, the impact of the farther nodes is quite limited, while the change mainly results from the nodes on the margin of the unit board.
For CVRP, the remaining solution percent decreases mainly because the change of the solution on the margin would result in the change of remaining capacity in the tour, which further leads to the change of whole subtour.
However, during the inference procedure of autoregressive solvers, since the agent could observe the change of remaining capacity and dynamically adjust the subtour, the farther nodes would have a lower impact than demonstrated in \cref{fig:change}.

\subsection{Detailed Experimental Settings} \label{sec:exp_d}

\begin{figure}[htbp]
    \centering
    \includegraphics[width=0.99\linewidth]{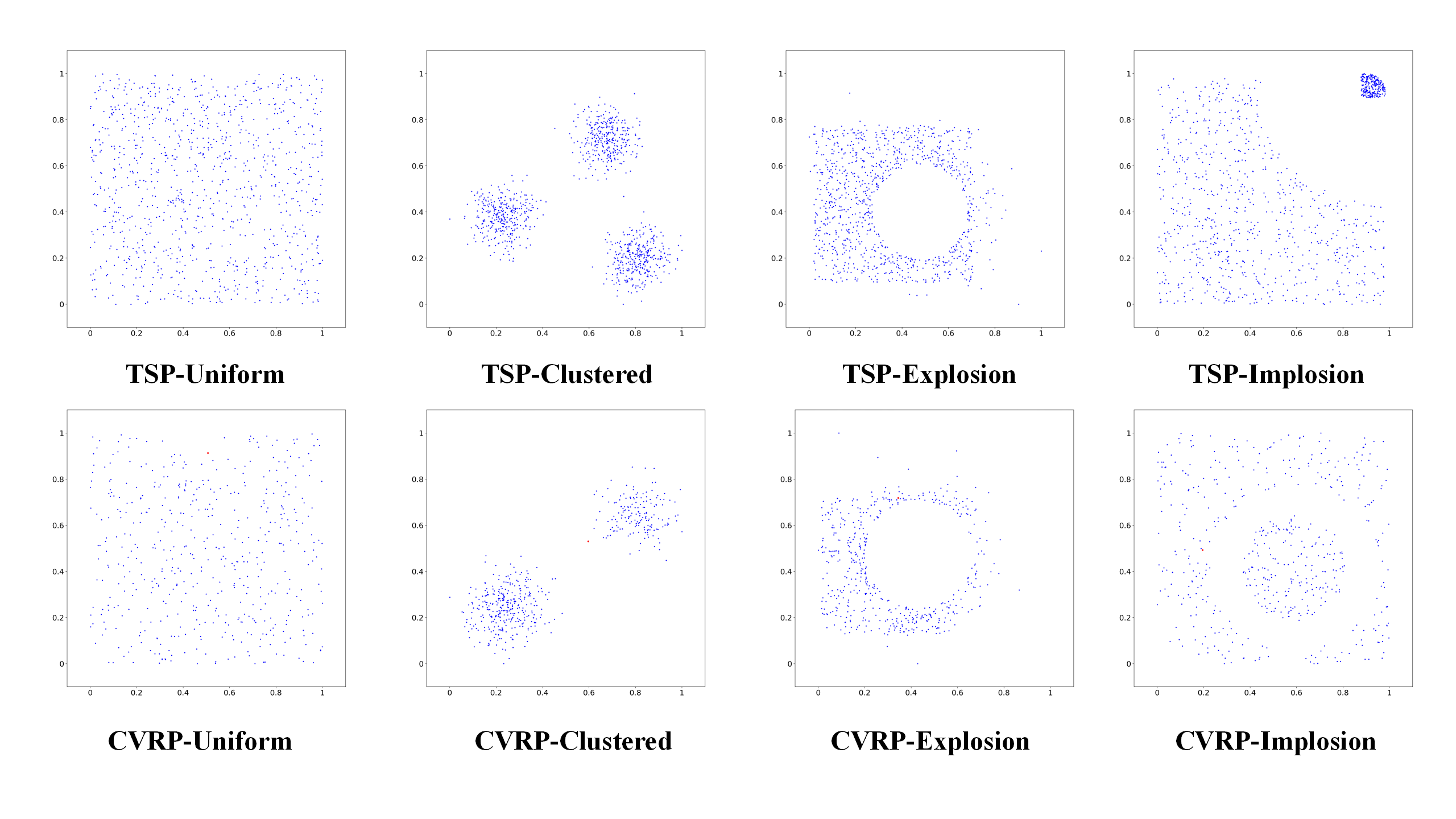}
    \caption{Sample Instances of the MSVDRP Dataset.
             Each sub-figure represents an instance in the dataset following its specified distribution from TSP1000/CVRP500.
             }
    \label{fig:sample}
\end{figure}

\paragraph{The MSVDRP Datasets.}

In our MSVDRP datasets, we include four different node distributions. 

Nodes for uniform distribution are uniformly generated from a $[0, 1]^2$ board.

Nodes for clustered distribution are generated as follows.
We first generate $N$ clusters on a $[0, L]$ following a uniform distribution.
Each node first selects its center uniformly, with additional Gaussian noise on coordinates with mean $\mu=1$ and standard deviation $\sigma = 0$. Our datasets include a balanced mixture of data with $N=3, L=10$ and $N=7, L = 50$.
For CVRP, the depot is generated uniformly together with the cluster.

Nodes for explosion distribution are first generated as uniform distribution.
Then a disc is determined by uniformly selecting a center on the board with a random radius uniformly selected from $[r_{\min}, r_{\max}]$.
All nodes inside this disc are mutated outside, following an exponential distribution with rate $\lambda$.
We choose $r_{\min} = 0.1, r_{\max}=0.5, \lambda=10$.

Nodes for implosion distribution are first generated as uniform distribution.
Then a disc is determined by uniformly selecting a center on the board with a random radius uniformly selected from $[r_{\min}, r_{\max}]$.
All nodes inside this disc are mutated closer to the center, following a multiplier $\lambda \in [1, +\infty)$.
The multiplier is determined by a truncated normal distribution with mean $\mu = 1$ and standard deviation $\sigma = 0$.
We choose $r_{\min} = 0.1, r_{\max}=0.5$.

For CVRP following uniform, explosion, and implosion distributions, the depot is generated uniformly together with the nodes; for CVRP following clustered distribution, the depot is generated together with the cluster centers.
Additionally, capacity is set to 50, and demands for each node is an integer randomly selected from 1 to 10.
All instances are scaled to $[0, 1]^2$ board after generation, which is shown in \cref{fig:sample}.
We refer to \citet{Bossek_Kerschke_Neumann_Wagner_Neumann_Trautmann_2019} for more technical details.

\paragraph{Solutions for Gap Calculation.}
Noted in \cref{sec:eval}, the (near-)optimal solutions for gap calculation are generated by various heuristic algorithms.
Their parameters are presented here.

TSP-100 is solved by Gurobi \citep{gurobi}, an exact solver, so the solutions are guaranteed to be truly optimal.
However, it fails on the other two large datasets due to unacceptable time consumption.
Instead, we use a SOTA heuristic algorithm LKH3 \citep{helsgaun_general_2009,helsgaun2017extension} for TSP-1000 and TSP-10000.
By controlling a reasonable time consumption, TSP-1000 is solved by LKH3 with 20000 iterations over 10 runs, whereas TSP-10000 is solved by LKH3 with 20000 iterations over 1 run.

For CVRP, we use a recently developed heuristic algorithm called HGS \citep{Vidal_2022}.
CVRP-50 and CVRP-500 are solved by HGS following the default parameters: 20000 iterations.
For CVRP-5000 solved by HGS, a 4-hour time limit is set for each instance, tolerating far fewer iterations than 20000.
This can partially explain why our method achieves better average gaps on CVRP-5000 datasets than on CVRP-500 datasets.
Again, we remark that comparisons among neural constructive methods are practically not affected by the quality of these solutions, so the sub-optimality of these heuristic algorithms is acceptable.

\paragraph{Implementation Settings}
Details in implementation are presented here, if not included in the main article.

For INViT, each single-view encoder contains 2 attention layers and the decoder contains 3 attention layers. The number of dimensions for features is 128 and the number of dimensions for feed forward layers is 512. The number of heads for each multi-head attention layer is 8. The default augmentation size is 8 and the default batch size is 64. For the whole training procedure, we train 500 epochs with 300 steps for each epoch. The default learning rate is $10^{-4}$.

For all evaluated methods, we keep the Pytorch version 1.12 on Python 3.9.
From our evaluations, the changes on Pytorch and Python versions do not cause any incompatibility issues for baseline methods.
In addition, the maximum split size of Pytorch GPU memory fragmentation is set to 512MB, to deal with potential memory issues.
In a few cases where those baseline methods still face CUDA out-of-memory issues, we reduce the pomo-size (i.e. number of generated solutions) to fit the memory budget of our machine. 

As previously mentioned, the pre-trained models from baseline methods are selected according to training size (small size around 100), but there are some more specifications.
For ELG-v1 on CVRP, there is no model trained on small-scale instances, so we re-train their model using their source code under a training size of 100, with local size k set to 50, borrowed from their TSP Model-100.
For Omni-TSP/CVRP, several models trained by variants of the algorithm are given.
We choose the FOMAML (First-Order Model-Agnostic Meta-Learning) version, which gives the best results compared with others in most of the cases.

Each instance, including TSPLIB and CVRPLIB, is scaled to $[0, 1]^2$ before model inference except for PointerFormer.
For PointerFormer, the instance is further scaled to $[0.05, 0.95]^2$ (for the MSVDRP dataset) or to $[0.1, 0.9]^2$ (for TSPLIB and CVRPLIB dataset) following their source code.
Using the unified input scale in PointerFormer reports unexpected runtime errors due to unknown reasons.

\subsection{Additional Discussion on Related Work} \label{sec:add_rw}

This section supplements more discussions on routing problem solvers in the realm of neural constructive methods.
We do not include a detailed discussion of these papers in our main article, in that they work in different but compatible directions from ours, and most of them are not considering similar generalization settings to ours.

\paragraph{Advanced Decoding Strategy.}
Inspired by POMO \citep{kwon_pomo_2020}, which generates multiple solutions in decoding steps to improve the model performance, many papers are devoted to exploring advanced decoding strategies to generate improved solutions by previous decoded solutions. 
This includes papers like AS \citep{Bello_Pham_Le_Norouzi_Bengio_2017}, MDAM \citep{Xin_Song_Cao_Zhang_2021}, SGBS \citep{Choo_Kwon_Kim_Jae_Hottung_Tierney_Gwon_2022}, EAS \citep{Hottung_Kwon_Tierney_2021}, LEHD \citep{NEURIPS2023_LEHD}, etc.
However, these methods make the decoding procedures no longer constructive, but iterative instead, resulting in much longer inference time to have enough solution improvements.

\paragraph{Divide and Conquer.}
Some papers partition the routing problem instance into multiple small instances and merge the partial solutions. 
An upper-level model is learned to partition the instance, and a lower-level model/solver is developed to construct partial solutions.
This includes papers like H-TSP \citep{Pan_Jin_Ding_Feng_Zhao_Song_Bian_2023}, TAM \citep{Hou_Yang_Su_Wang_Deng_2022}, GLOP \citep{Ye_Wang_Liang_Cao_Li_Li_2023}, etc. 
However, due to the existence of an upper-level model to segment the full instance, their methods need to include much larger training instances.

\paragraph{Local Search Combination.}
For TSP, many papers explore a direct combination between neural constructive models and local search algorithms.
A popular local search algorithm is the MCTS-framework algorithm \citep{fu_generalize_2021}.
This includes papers like GPN \citep{ma_combinatorial_2019} with 2-opt, Att-GCRN \citep{fu_generalize_2021} with MCTS, DIMES \citep{Qiu_Sun_Yang_2022} with MCTS, DIFFUSCO \citep{sun_difusco_2023} with MCTS, etc.
Different from learn-to-search methods which interleave the neural network models and the search algorithms, these papers develop constructive models instead and apply local search at the end.

\subsection{TSPLIB and CVRPLIB Results} \label{sec:lib_d}

The results for each TSPLIB and CVRPLIB instance of baseline methods and our method are displayed in \cref{table:tsplib_d} and \cref{table:cvrplib_d} respectively.

\centering

\begin{longtable}{ccccccccc}
\caption{Detailed Results for all included TSPLIB instances.} \label{table:tsplib_d} \\

\toprule
Instance & \textbf{\methodname{}-2} & \textbf{\methodname{}-3} & BQ-NCO & LEHD & ELG & Omni-TSP & POMO & PointerFormer \\
\midrule
\endfirsthead

\multicolumn{9}{c}{{\bfseries \tablename\ \thetable{} -- continued from previous page}} \\
\toprule
Instance & \textbf{\methodname{}-2} & \textbf{\methodname{}-3} & BQ-NCO & LEHD & ELG & Omni-TSP & POMO & PointerFormer \\
\midrule
\endhead

\bottomrule
\multicolumn{9}{r}{{Continued on next page}} \\
\endfoot

\bottomrule
\endlastfoot

eil51 & $1.64$\% & $1.17$\% & $2.82$\% & $1.64$\% & $\bm{0.94}\%$ & $1.17$\% & $1.41$\% & $\bm{0.94}\%$ \\
berlin52 & $1.41$\% & $\bm{0.04}\%$ & $17.09$\% & $\bm{0.04}\%$ & $0.11$\% & $6.15$\% & $\bm{0.04}\%$ & $\bm{0.04}\%$ \\
st70 & $1.63$\% & $0.89$\% & $2.07$\% & $\bm{0.44}\%$ & $\bm{0.44}\%$ & $1.63$\% & $\bm{0.44}\%$ & $\bm{0.44}\%$ \\
pr76 & $1.28$\% & $1.28$\% & $0.11$\% & $0.22$\% & $0.82$\% & $0.82$\% & $\bm{0.00}\%$ & $0.14$\% \\
eil76 & $2.79$\% & $\bm{0.74}\%$ & $5.02$\% & $2.60$\% & $2.23$\% & $4.28$\% & $1.30$\% & $2.79$\% \\
rat99 & $3.39$\% & $1.82$\% & $18.50$\% & $\bm{1.16}\%$ & $2.40$\% & $2.48$\% & $7.18$\% & $4.87$\% \\
kroA100 & $0.84$\% & $1.15$\% & $12.15$\% & $\bm{0.12}\%$ & $1.72$\% & $\bm{0.12}\%$ & $3.27$\% & $1.95$\% \\
kroE100 & $1.75$\% & $1.68$\% & $13.63$\% & $\bm{0.43}\%$ & $1.01$\% & $0.79$\% & $3.17$\% & $0.95$\% \\
kroB100 & $1.53$\% & $0.79$\% & $4.35$\% & $\bm{0.26}\%$ & $1.35$\% & $2.43$\% & $2.81$\% & $1.87$\% \\
rd100 & $2.93$\% & $1.30$\% & $9.51$\% & $\bm{0.01}\%$ & $0.06$\% & $0.63$\% & $\bm{0.01}\%$ & $0.13$\% \\
kroD100 & $2.34$\% & $1.53$\% & $11.13$\% & $\bm{0.39}\%$ & $1.53$\% & $2.52$\% & $2.00$\% & $1.10$\% \\
kroC100 & $0.55$\% & $1.27$\% & $7.50$\% & $\bm{0.33}\%$ & $1.15$\% & $0.74$\% & $1.39$\% & $1.16$\% \\
eil101 & $5.88$\% & $2.70$\% & $4.77$\% & $2.38$\% & $2.70$\% & $3.02$\% & $\bm{2.07}\%$ & $\bm{2.07}\%$ \\
lin105 & $2.23$\% & $1.79$\% & $12.35$\% & $\bm{0.35}\%$ & $3.00$\% & $3.32$\% & $1.43$\% & $2.87$\% \\
pr107 & $\bm{0.57}\%$ & $1.54$\% & $13.74$\% & $11.82$\% & $2.61$\% & $1.14$\% & $2.99$\% & $3.37$\% \\
pr124 & $1.42$\% & $1.23$\% & $16.84$\% & $1.11$\% & $1.02$\% & $1.28$\% & $0.70$\% & $\bm{0.27}\%$ \\
bier127 & $3.24$\% & $3.24$\% & $6.29$\% & $5.63$\% & $16.29$\% & $\bm{3.01}\%$ & $5.81$\% & $14.96$\% \\
ch130 & $1.77$\% & $2.00$\% & $\bm{0.21}\%$ & $0.56$\% & $1.75$\% & $2.36$\% & $0.62$\% & $0.41$\% \\
pr136 & $2.29$\% & $1.43$\% & $9.87$\% & $\bm{0.45}\%$ & $1.73$\% & $0.77$\% & $1.04$\% & $1.14$\% \\
pr144 & $2.83$\% & $2.34$\% & $14.73$\% & $3.59$\% & $0.52$\% & $1.86$\% & $1.88$\% & $\bm{0.37}\%$ \\
kroA150 & $1.67$\% & $2.43$\% & $4.95$\% & $\bm{1.40}\%$ & $2.38$\% & $1.84$\% & $3.60$\% & $6.59$\% \\
kroB150 & $4.02$\% & $2.57$\% & $7.19$\% & $\bm{0.76}\%$ & $1.42$\% & $1.44$\% & $2.96$\% & $2.55$\% \\
ch150 & $2.79$\% & $2.40$\% & $5.65$\% & $\bm{0.60}\%$ & $1.67$\% & $1.55$\% & $0.69$\% & $0.78$\% \\
pr152 & $5.64$\% & $7.97$\% & $11.92$\% & $12.14$\% & $1.81$\% & $1.40$\% & $1.69$\% & $\bm{0.45}\%$ \\
u159 & $1.71$\% & $1.01$\% & $\bm{0.00}\%$ & $1.45$\% & $1.32$\% & $1.20$\% & $1.35$\% & $1.05$\% \\
rat195 & $3.10$\% & $3.36$\% & $10.93$\% & $\bm{1.42}\%$ & $7.32$\% & $6.29$\% & $10.16$\% & $13.52$\% \\
d198 & $10.07$\% & $8.09$\% & $10.32$\% & $9.24$\% & $27.05$\% & $\bm{3.54}\%$ & $32.84$\% & $20.21$\% \\
kroA200 & $2.80$\% & $2.51$\% & $8.79$\% & $\bm{0.64}\%$ & $3.92$\% & $1.77$\% & $3.78$\% & $8.90$\% \\
kroB200 & $2.34$\% & $3.98$\% & $10.74$\% & $\bm{0.16}\%$ & $3.58$\% & $1.63$\% & $4.39$\% & $8.56$\% \\
tsp225 & $1.79$\% & $1.05$\% & $4.70$\% & $\bm{0.69}\%$ & $4.60$\% & $3.63$\% & $7.69$\% & $11.62$\% \\
ts225 & $3.90$\% & $3.12$\% & $13.48$\% & $\bm{0.28}\%$ & $5.57$\% & $2.53$\% & $8.48$\% & $2.65$\% \\
pr226 & $8.74$\% & $4.24$\% & $11.76$\% & $\bm{0.87}\%$ & $1.03$\% & $1.95$\% & $4.56$\% & $2.09$\% \\
gil262 & $4.16$\% & $5.09$\% & $4.79$\% & $\bm{1.60}\%$ & $3.20$\% & $3.36$\% & $4.16$\% & $2.61$\% \\
pr264 & $\bm{1.95}\%$ & $2.02$\% & $12.50$\% & $5.17$\% & $4.03$\% & $2.67$\% & $16.02$\% & $10.21$\% \\
a280 & $5.27$\% & $6.79$\% & $\bm{0.47}\%$ & $3.02$\% & $8.06$\% & $5.54$\% & $13.77$\% & $13.18$\% \\
pr299 & $5.74$\% & $5.29$\% & $6.65$\% & $\bm{2.91}\%$ & $5.84$\% & $4.43$\% & $15.33$\% & $17.58$\% \\
lin318 & $5.42$\% & $3.49$\% & $10.36$\% & $\bm{1.41}\%$ & $6.21$\% & $4.84$\% & $12.34$\% & $8.54$\% \\
rd400 & $4.17$\% & $5.37$\% & $3.06$\% & $\bm{1.01}\%$ & $6.39$\% & $5.24$\% & $14.70$\% & $10.89$\% \\
fl417 & $8.61$\% & $8.57$\% & $19.02$\% & $\bm{4.78}\%$ & $8.30$\% & $7.57$\% & $15.47$\% & $7.23$\% \\
pr439 & $7.12$\% & $7.42$\% & $7.14$\% & $\bm{3.37}\%$ & $6.97$\% & $5.61$\% & $24.08$\% & $18.02$\% \\
pcb442 & $2.79$\% & $4.44$\% & $\bm{0.90}\%$ & $3.11$\% & $10.08$\% & $7.78$\% & $18.26$\% & $16.57$\% \\
d493 & $9.33$\% & $\bm{5.80}\%$ & $8.00$\% & $9.49$\% & $58.60$\% & $9.79$\% & $76.74$\% & $35.85$\% \\
u574 & $6.88$\% & $6.02$\% & $\bm{1.76}\%$ & $2.73$\% & $10.05$\% & $10.25$\% & $26.05$\% & $23.08$\% \\
rat575 & $5.76$\% & $5.96$\% & $10.08$\% & $\bm{3.03}\%$ & $8.53$\% & $14.07$\% & $26.16$\% & $24.23$\% \\
p654 & $15.26$\% & $\bm{9.31}\%$ & $16.03$\% & $10.24$\% & $9.78$\% & $10.67$\% & $28.73$\% & $14.14$\% \\
d657 & $9.52$\% & $\bm{7.52}\%$ & $8.62$\% & $8.05$\% & $18.43$\% & $12.32$\% & $34.53$\% & $27.21$\% \\
u724 & $5.76$\% & $5.30$\% & $\bm{2.18}\%$ & $3.27$\% & $11.25$\% & $15.22$\% & $27.16$\% & $22.82$\% \\
rat783 & $5.20$\% & $5.74$\% & $9.81$\% & $\bm{4.28}\%$ & $10.88$\% & $17.42$\% & $33.26$\% & $26.25$\% \\
pr1002 & $9.38$\% & $11.56$\% & $8.74$\% & $\bm{4.44}\%$ & $11.25$\% & $18.44$\% & $40.16$\% & $25.50$\% \\
u1060 & $9.40$\% & $10.38$\% & $\bm{8.63}\%$ & $8.92$\% & $12.22$\% & $21.98$\% & $46.47$\% & $33.86$\% \\
vm1084 & $7.42$\% & $6.72$\% & $10.38$\% & $\bm{5.98}\%$ & $19.43$\% & $19.04$\% & $43.18$\% & $30.27$\% \\
pcb1173 & $\bm{6.04}\%$ & $7.24$\% & $11.71$\% & $6.34$\% & $15.68$\% & $24.68$\% & $44.42$\% & $31.15$\% \\
d1291 & $11.36$\% & $\bm{10.87}\%$ & $11.14$\% & $14.13$\% & $12.21$\% & $27.26$\% & $145.87$\% & $26.52$\% \\
rl1304 & $8.87$\% & $8.77$\% & $8.77$\% & $\bm{7.75}\%$ & $15.25$\% & $26.93$\% & $56.70$\% & $32.11$\% \\
rl1323 & $8.90$\% & $8.62$\% & $\bm{7.63}\%$ & $9.26$\% & $16.79$\% & $26.07$\% & $55.96$\% & $32.69$\% \\
nrw1379 & $6.30$\% & $\bm{6.03}\%$ & $9.83$\% & $9.91$\% & $12.18$\% & $21.24$\% & $36.06$\% & $27.62$\% \\
fl1400 & $12.26$\% & $\bm{12.24}\%$ & $31.19$\% & $18.80$\% & $30.61$\% & $16.39$\% & $40.15$\% & $18.76$\% \\
u1432 & $5.30$\% & $4.81$\% & $4.98$\% & $\bm{3.34}\%$ & $11.62$\% & $19.43$\% & $31.40$\% & $21.53$\% \\
fl1577 & $9.15$\% & $\bm{8.49}\%$ & $21.61$\% & $17.63$\% & $14.19$\% & $29.08$\% & $74.78$\% & $25.32$\% \\
d1655 & $\bm{12.32}\%$ & $12.84$\% & $17.01$\% & $13.89$\% & $19.24$\% & $30.77$\% & $58.94$\% & $34.38$\% \\
vm1748 & $\bm{7.92}\%$ & $9.46$\% & $11.18$\% & $14.81$\% & $19.28$\% & $26.67$\% & $53.97$\% & $31.73$\% \\
u1817 & $\bm{6.66}\%$ & $8.40$\% & $9.43$\% & $10.28$\% & $15.62$\% & $35.97$\% & $59.63$\% & $34.44$\% \\
rl1889 & $11.78$\% & $\bm{9.42}\%$ & $14.91$\% & $10.64$\% & $18.54$\% & $33.30$\% & $67.05$\% & $35.20$\% \\
d2103 & $9.75$\% & $\bm{7.53}\%$ & $17.47$\% & $14.57$\% & $15.10$\% & $33.54$\% & $64.58$\% & $22.21$\% \\
u2152 & $8.59$\% & $\bm{7.87}\%$ & $9.08$\% & $11.62$\% & $17.58$\% & $37.83$\% & $59.57$\% & $37.22$\% \\
u2319 & $\bm{0.98}\%$ & $1.41$\% & $3.41$\% & $2.54$\% & $4.02$\% & $16.73$\% & $23.53$\% & $12.41$\% \\
pr2392 & $9.00$\% & $\bm{8.75}\%$ & $9.26$\% & $10.98$\% & $18.23$\% & $38.31$\% & $60.63$\% & $38.57$\% \\
pcb3038 & $8.23$\% & $\bm{7.29}\%$ & $13.44$\% & $13.04$\% & $17.53$\% & $38.11$\% & $59.24$\% & $36.91$\% \\
fl3795 & $13.20$\% & $\bm{10.74}\%$ & $32.09$\% & $13.55$\% & $22.15$\% & $42.58$\% & $92.85$\% & $15.31$\% \\
fnl4461 & $6.77$\% & $\bm{6.64}\%$ & $21.38$\% & $18.79$\% & $17.42$\% & $41.94$\% & $59.62$\% & $34.14$\% \\
rl5915 & $\bm{9.50}\%$ & $11.64$\% & $24.58$\% & $22.34$\% & $23.06$\% & $64.59$\% & $83.28$\% & $48.38$\% \\
rl5934 & $12.34$\% & $\bm{8.97}\%$ & $30.17$\% & $35.47$\% & $23.88$\% & $65.84$\% & $83.16$\% & $42.81$\% \\
rl11849 & $10.38$\% & $\bm{9.91}\%$ & $45.21$\% & $41.69$\% & $24.01$\% & $77.16$\% & $92.57$\% & $52.02$\% \\
usa13509 & $12.19$\% & $\bm{10.29}\%$ & OOM & $55.80$\% & $32.59$\% & $107.88$\% & $116.26$\% & $67.06$\% \\
brd14051 & $9.41$\% & $\bm{8.82}\%$ & OOM & $42.54$\% & $25.70$\% & $78.22$\% & $91.32$\% & $49.20$\% \\
d15112 & $\bm{7.84}\%$ & $8.45$\% & OOM & $34.41$\% & $25.69$\% & $73.40$\% & $82.63$\% & $46.62$\% \\
d18512 & $\bm{7.87}\%$ & $8.07$\% & OOM & OOM & $25.65$\% & $77.97$\% & $88.59$\% & $47.41$\% \\

\end{longtable}

\begin{longtable}{ccccccccc}
\caption{Detailed Results for all included CVRPLIB Set-X instances.} \label{table:cvrplib_d} \\

\toprule
Instance & \textbf{\methodname{}-2} & \textbf{\methodname{}-3} & BQ-NCO & LEHD & ELG & Omni-TSP & POMO \\
\midrule
\endfirsthead

\multicolumn{8}{c}{{\bfseries \tablename\ \thetable{} -- continued from previous page}} \\
\toprule
Instance & \textbf{\methodname{}-2} & \textbf{\methodname{}-3} & BQ-NCO & LEHD & ELG & Omni-TSP & POMO \\
\midrule
\endhead

\bottomrule
\multicolumn{8}{r}{{Continued on next page}} \\
\endfoot

\bottomrule
\endlastfoot

X-n101-k25 & $3.70$\% & $\bm{2.61}\%$ & $21.84$\% & $13.97$\% & $3.66$\% & $5.80$\% & $9.01$\% \\
X-n106-k14 & $4.17$\% & $4.75$\% & $7.05$\% & $3.75$\% & $\bm{3.01}\%$ & $7.59$\% & $4.64$\% \\
X-n110-k13 & $7.03$\% & $5.06$\% & $4.99$\% & $\bm{1.88}\%$ & $8.34$\% & $3.49$\% & $2.85$\% \\
X-n115-k10 & $5.38$\% & $7.82$\% & $19.76$\% & $9.44$\% & $\bm{3.96}\%$ & $12.19$\% & $15.49$\% \\
X-n120-k6 & $8.18$\% & $7.73$\% & $16.44$\% & $\bm{3.89}\%$ & $7.36$\% & $5.69$\% & $12.16$\% \\
X-n125-k30 & $\bm{3.08}\%$ & $7.18$\% & $13.11$\% & $17.89$\% & $4.42$\% & $9.44$\% & $5.31$\% \\
X-n129-k18 & $7.40$\% & $7.25$\% & $5.96$\% & $4.01$\% & $2.65$\% & $6.00$\% & $\bm{2.18}\%$ \\
X-n134-k13 & $5.88$\% & $\bm{5.57}\%$ & $9.83$\% & $8.94$\% & $5.84$\% & $6.07$\% & $5.77$\% \\
X-n139-k10 & $4.75$\% & $5.92$\% & $9.22$\% & $\bm{3.08}\%$ & $8.91$\% & $4.75$\% & $4.16$\% \\
X-n143-k7 & $7.26$\% & $8.46$\% & $13.41$\% & $14.01$\% & $7.25$\% & $8.87$\% & $\bm{4.43}\%$ \\
X-n148-k46 & $7.57$\% & $4.36$\% & $10.05$\% & $38.98$\% & $\bm{2.81}\%$ & $9.12$\% & $10.86$\% \\
X-n153-k22 & $13.22$\% & $11.62$\% & $34.22$\% & $28.93$\% & $\bm{11.10}\%$ & $15.27$\% & $19.02$\% \\
X-n157-k13 & $12.57$\% & $5.92$\% & $9.60$\% & $4.72$\% & $11.15$\% & $\bm{4.50}\%$ & $21.81$\% \\
X-n162-k11 & $5.74$\% & $5.80$\% & $9.46$\% & $\bm{3.47}\%$ & $8.38$\% & $7.35$\% & $7.72$\% \\
X-n167-k10 & $7.41$\% & $5.51$\% & $13.51$\% & $\bm{5.03}\%$ & $13.94$\% & $5.33$\% & $6.38$\% \\
X-n172-k51 & $9.25$\% & $8.59$\% & $14.88$\% & $33.28$\% & $\bm{4.26}\%$ & $7.40$\% & $15.01$\% \\
X-n176-k26 & $14.97$\% & $\bm{4.85}\%$ & $20.57$\% & $27.00$\% & $7.75$\% & $10.90$\% & $11.71$\% \\
X-n181-k23 & $14.40$\% & $8.24$\% & $5.80$\% & $\bm{1.44}\%$ & $5.98$\% & $6.46$\% & $8.18$\% \\
X-n186-k15 & $6.17$\% & $6.54$\% & $9.42$\% & $\bm{3.64}\%$ & $12.64$\% & $7.07$\% & $7.31$\% \\
X-n190-k8 & $8.24$\% & $6.37$\% & $9.71$\% & $\bm{5.41}\%$ & $7.33$\% & $10.29$\% & $8.38$\% \\
X-n195-k51 & $8.98$\% & $9.73$\% & $22.75$\% & $15.47$\% & $\bm{3.60}\%$ & $10.84$\% & $15.78$\% \\
X-n200-k36 & $5.72$\% & $5.62$\% & $12.13$\% & $10.26$\% & $\bm{4.54}\%$ & $7.16$\% & $9.37$\% \\
X-n204-k19 & $7.69$\% & $9.32$\% & $13.53$\% & $\bm{2.69}\%$ & $14.19$\% & $8.26$\% & $10.68$\% \\
X-n209-k16 & $6.68$\% & $7.92$\% & $9.63$\% & $\bm{3.88}\%$ & $11.07$\% & $5.08$\% & $8.18$\% \\
X-n214-k11 & $12.55$\% & $11.84$\% & $12.56$\% & $\bm{7.21}\%$ & $7.88$\% & $8.60$\% & $8.94$\% \\
X-n219-k73 & $9.46$\% & $2.36$\% & $9.03$\% & $15.51$\% & $\bm{1.27}\%$ & $3.16$\% & $3.92$\% \\
X-n223-k34 & $8.20$\% & $8.28$\% & $7.95$\% & $5.80$\% & $\bm{4.73}\%$ & $7.14$\% & $9.66$\% \\
X-n228-k23 & $12.12$\% & $8.31$\% & $23.02$\% & $14.41$\% & $\bm{8.12}\%$ & $10.19$\% & $18.73$\% \\
X-n233-k16 & $10.85$\% & $11.82$\% & $10.17$\% & $\bm{5.64}\%$ & $7.07$\% & $7.31$\% & $11.72$\% \\
X-n237-k14 & $10.34$\% & $12.07$\% & $\bm{3.46}\%$ & $4.06$\% & $13.97$\% & $5.05$\% & $25.80$\% \\
X-n242-k48 & $4.48$\% & $2.75$\% & $5.75$\% & $6.02$\% & $\bm{2.69}\%$ & $5.99$\% & $8.36$\% \\
X-n247-k50 & $15.14$\% & $14.15$\% & $27.82$\% & $44.92$\% & $\bm{9.37}\%$ & $9.46$\% & $18.40$\% \\
X-n251-k28 & $7.45$\% & $7.40$\% & $4.56$\% & $\bm{2.37}\%$ & $8.67$\% & $6.33$\% & $11.08$\% \\
X-n256-k16 & $8.15$\% & $\bm{6.15}\%$ & $14.22$\% & $7.30$\% & $13.83$\% & $8.16$\% & $20.04$\% \\
X-n261-k13 & $8.82$\% & $9.33$\% & $8.85$\% & $8.83$\% & $9.35$\% & $\bm{6.61}\%$ & $11.22$\% \\
X-n266-k58 & $8.41$\% & $8.02$\% & $7.02$\% & $6.38$\% & $\bm{5.54}\%$ & $8.96$\% & $12.64$\% \\
X-n270-k35 & $8.88$\% & $9.36$\% & $\bm{5.71}\%$ & $6.54$\% & $13.13$\% & $8.19$\% & $13.13$\% \\
X-n275-k28 & $14.70$\% & $14.37$\% & $\bm{4.96}\%$ & $6.58$\% & $10.60$\% & $7.27$\% & $18.19$\% \\
X-n280-k17 & $9.81$\% & $9.03$\% & $12.04$\% & $9.64$\% & $7.31$\% & $\bm{5.48}\%$ & $11.66$\% \\
X-n284-k15 & $9.93$\% & $9.71$\% & $9.95$\% & $\bm{4.99}\%$ & $10.54$\% & $18.46$\% & $13.43$\% \\
X-n289-k60 & $7.34$\% & $7.08$\% & $8.19$\% & $9.90$\% & $\bm{4.06}\%$ & $8.26$\% & $11.14$\% \\
X-n294-k50 & $9.39$\% & $\bm{4.67}\%$ & $10.99$\% & $9.68$\% & $5.20$\% & $10.75$\% & $17.70$\% \\
X-n298-k31 & $9.03$\% & $11.08$\% & $7.34$\% & $7.65$\% & $8.71$\% & $\bm{7.14}\%$ & $16.70$\% \\
X-n303-k21 & $7.91$\% & $9.25$\% & $8.94$\% & $\bm{2.83}\%$ & $6.31$\% & $6.73$\% & $15.92$\% \\
X-n308-k13 & $10.33$\% & $10.63$\% & $9.53$\% & $\bm{4.49}\%$ & $11.34$\% & $8.42$\% & $17.34$\% \\
X-n313-k71 & $8.01$\% & $7.52$\% & $12.05$\% & $14.06$\% & $\bm{3.77}\%$ & $7.82$\% & $12.58$\% \\
X-n317-k53 & $8.96$\% & $7.67$\% & $5.18$\% & $\bm{3.88}\%$ & $5.02$\% & $4.25$\% & $69.93$\% \\
X-n322-k28 & $8.52$\% & $9.80$\% & $8.11$\% & $\bm{3.89}\%$ & $14.94$\% & $6.26$\% & $19.26$\% \\
X-n327-k20 & $9.78$\% & $8.05$\% & $8.63$\% & $10.74$\% & $14.66$\% & $\bm{7.25}\%$ & $22.13$\% \\
X-n331-k15 & $10.21$\% & $10.50$\% & $8.59$\% & $\bm{3.36}\%$ & $14.63$\% & $5.96$\% & $59.53$\% \\
X-n336-k84 & $8.98$\% & $8.38$\% & $13.66$\% & $18.78$\% & $\bm{3.70}\%$ & $8.37$\% & $13.60$\% \\
X-n344-k43 & $10.41$\% & $10.35$\% & $7.57$\% & $\bm{3.93}\%$ & $10.78$\% & $8.94$\% & $20.30$\% \\
X-n351-k40 & $10.14$\% & $9.59$\% & $13.02$\% & $\bm{7.48}\%$ & $8.35$\% & $12.43$\% & $23.99$\% \\
X-n359-k29 & $8.83$\% & $8.27$\% & $6.21$\% & $\bm{2.79}\%$ & $6.00$\% & $5.02$\% & $10.38$\% \\
X-n367-k17 & $11.38$\% & $9.95$\% & $11.15$\% & $\bm{9.44}\%$ & $9.80$\% & $10.98$\% & $19.40$\% \\
X-n376-k94 & $5.78$\% & $6.11$\% & $5.13$\% & $5.45$\% & $\bm{1.96}\%$ & $3.54$\% & $19.79$\% \\
X-n384-k52 & $\bm{6.29}\%$ & $7.21$\% & $6.93$\% & $7.79$\% & $10.19$\% & $7.50$\% & $17.74$\% \\
X-n393-k38 & $10.32$\% & $9.30$\% & $11.75$\% & $\bm{4.47}\%$ & $13.62$\% & $9.85$\% & $24.97$\% \\
X-n401-k29 & $7.44$\% & $6.28$\% & $9.88$\% & $\bm{5.55}\%$ & $10.48$\% & $5.60$\% & $13.21$\% \\
X-n411-k19 & $15.84$\% & $14.91$\% & $\bm{10.61}\%$ & $13.49$\% & $15.67$\% & $14.29$\% & $40.13$\% \\
X-n420-k130 & $16.78$\% & $16.25$\% & $20.61$\% & $61.64$\% & $\bm{4.24}\%$ & $14.40$\% & $22.21$\% \\
X-n429-k61 & $7.71$\% & $7.38$\% & $8.36$\% & $\bm{6.25}\%$ & $11.24$\% & $8.65$\% & $30.63$\% \\
X-n439-k37 & $14.92$\% & $14.95$\% & $5.39$\% & $\bm{2.00}\%$ & $11.91$\% & $8.05$\% & $31.51$\% \\
X-n449-k29 & $7.61$\% & $7.52$\% & $14.47$\% & $\bm{7.18}\%$ & $9.49$\% & $7.44$\% & $25.14$\% \\
X-n459-k26 & $11.04$\% & $11.37$\% & $11.74$\% & $\bm{8.66}\%$ & $14.05$\% & $10.78$\% & $30.76$\% \\
X-n469-k138 & $7.03$\% & $7.14$\% & $14.26$\% & $28.32$\% & $\bm{6.89}\%$ & $9.15$\% & $14.47$\% \\
X-n480-k70 & $8.54$\% & $7.55$\% & $5.78$\% & $\bm{3.06}\%$ & $8.16$\% & $7.43$\% & $24.92$\% \\
X-n491-k59 & $8.41$\% & $8.40$\% & $10.61$\% & $\bm{5.08}\%$ & $6.99$\% & $11.40$\% & $27.87$\% \\
X-n502-k39 & $12.40$\% & $11.05$\% & $5.27$\% & $\bm{3.31}\%$ & $27.77$\% & $7.30$\% & $56.62$\% \\
X-n513-k21 & $13.90$\% & $12.88$\% & $7.37$\% & $\bm{4.97}\%$ & $21.89$\% & $11.02$\% & $57.73$\% \\
X-n524-k153 & $14.33$\% & $14.20$\% & $29.80$\% & $78.66$\% & $\bm{9.23}\%$ & $13.42$\% & $28.64$\% \\
X-n536-k96 & $8.32$\% & $8.01$\% & $10.66$\% & $13.92$\% & $\bm{6.29}\%$ & $9.33$\% & $29.50$\% \\
X-n548-k50 & $13.81$\% & $14.19$\% & $\bm{2.50}\%$ & $3.32$\% & $10.03$\% & $6.42$\% & $30.66$\% \\
X-n561-k42 & $6.14$\% & $12.51$\% & $6.73$\% & $\bm{5.86}\%$ & $12.24$\% & $8.86$\% & $67.95$\% \\
X-n573-k30 & $8.29$\% & $\bm{6.53}\%$ & $13.45$\% & $11.35$\% & $11.60$\% & $18.10$\% & $41.22$\% \\
X-n586-k159 & $10.73$\% & $10.25$\% & $11.73$\% & $17.12$\% & $\bm{7.25}\%$ & $8.72$\% & $25.73$\% \\
X-n599-k92 & $7.39$\% & $7.46$\% & $\bm{6.17}\%$ & $8.93$\% & $8.61$\% & $9.22$\% & $29.18$\% \\
X-n613-k62 & $10.00$\% & $10.03$\% & $10.32$\% & $\bm{6.89}\%$ & $9.80$\% & $10.65$\% & $89.65$\% \\
X-n627-k43 & $8.50$\% & $8.33$\% & $9.72$\% & $\bm{5.50}\%$ & $16.78$\% & $8.78$\% & $61.36$\% \\
X-n641-k35 & $8.16$\% & $8.19$\% & $\bm{4.28}\%$ & $5.66$\% & $16.40$\% & $9.15$\% & $21.82$\% \\
X-n655-k131 & $6.05$\% & $10.18$\% & $4.54$\% & $5.24$\% & $\bm{4.23}\%$ & $5.43$\% & $23.02$\% \\
X-n670-k130 & $16.68$\% & $15.87$\% & $30.49$\% & $98.95$\% & $\bm{10.11}\%$ & $16.46$\% & $44.83$\% \\
X-n685-k75 & $11.21$\% & $11.35$\% & $13.75$\% & $\bm{8.34}\%$ & $9.38$\% & $11.29$\% & $78.15$\% \\
X-n701-k44 & $7.87$\% & $8.65$\% & $6.50$\% & $\bm{4.11}\%$ & $8.92$\% & $7.94$\% & $52.05$\% \\
X-n716-k35 & $9.78$\% & $\bm{8.19}\%$ & $14.26$\% & $8.80$\% & $14.97$\% & $12.50$\% & $44.14$\% \\
X-n733-k159 & $12.43$\% & $6.86$\% & $11.76$\% & $14.72$\% & $\bm{5.36}\%$ & $11.81$\% & $57.41$\% \\
X-n749-k98 & $8.42$\% & $9.41$\% & $11.55$\% & $8.97$\% & $\bm{7.67}\%$ & $14.73$\% & $37.53$\% \\
X-n766-k71 & $12.42$\% & $12.16$\% & $17.32$\% & $15.49$\% & $\bm{7.98}\%$ & $10.28$\% & $41.54$\% \\
X-n783-k48 & $9.29$\% & $7.94$\% & $12.99$\% & $\bm{5.64}\%$ & $16.36$\% & $15.49$\% & $105.04$\% \\
X-n801-k40 & $12.39$\% & $12.25$\% & $5.80$\% & $\bm{3.48}\%$ & $16.59$\% & $8.98$\% & $124.56$\% \\
X-n819-k171 & $8.54$\% & $\bm{7.53}\%$ & $11.51$\% & $12.21$\% & $8.35$\% & $11.50$\% & $86.87$\% \\
X-n837-k142 & $8.38$\% & $8.43$\% & $\bm{4.56}\%$ & $6.68$\% & $7.07$\% & $10.53$\% & $21.08$\% \\
X-n856-k95 & $17.80$\% & $18.55$\% & $3.65$\% & $\bm{2.97}\%$ & $11.56$\% & $8.87$\% & $87.82$\% \\
X-n876-k59 & $7.51$\% & $\bm{7.14}\%$ & $13.10$\% & $7.30$\% & $12.97$\% & $11.92$\% & $78.67$\% \\
X-n895-k37 & $10.29$\% & $9.24$\% & $\bm{7.91}\%$ & $8.70$\% & $21.53$\% & $14.18$\% & $56.53$\% \\
X-n916-k207 & $9.78$\% & $9.06$\% & $9.10$\% & $15.36$\% & $\bm{6.03}\%$ & $9.27$\% & $18.34$\% \\
X-n936-k151 & $\bm{7.93}\%$ & $8.19$\% & $35.15$\% & $139.42$\% & $12.86$\% & $20.60$\% & $57.97$\% \\
X-n957-k87 & $15.66$\% & $15.48$\% & $5.92$\% & $\bm{3.85}\%$ & $23.85$\% & $12.52$\% & $115.04$\% \\
X-n979-k58 & $8.13$\% & $\bm{7.35}\%$ & $13.16$\% & $19.79$\% & $11.52$\% & $7.59$\% & $69.18$\% \\
X-n1001-k43 & $9.74$\% & $9.35$\% & $\bm{6.87}\%$ & $7.57$\% & $15.65$\% & $15.87$\% & $142.29$\% \\

\end{longtable}



\end{document}